\documentclass[10pt,twocolumn,letterpaper]{article}

\usepackage[pagenumbers]{cvpr}

\usepackage{placeins}

\definecolor{cvprblue}{rgb}{0.21,0.49,0.74}
\usepackage[breaklinks,colorlinks,allcolors=cvprblue]{hyperref}

\hypersetup{
    pdftitle={Calibrating Track-Query Updates at Test Time for Multi-Object Tracking},
    pdfauthor={Riku Inoue, Shogo Sato, Yu Mitsuzumi, Kazuhiko Murasaki, Tomoyasu Shimada, Takashi Sano}
}

\title{Calibrating Track-Query Updates at Test Time for Multi-Object Tracking}

\author{
Riku Inoue, \enspace Shogo Sato, \enspace Yu Mitsuzumi\\
Kazuhiko Murasaki, \enspace Tomoyasu Shimada, \enspace Takashi Sano\\[0.07cm]
NTT, Inc.\\
{\tt\small riku.inoue@ntt.com}
}

\begin{document}

\maketitle


\begin{abstract}
Multi-Object Tracking (MOT) requires localizing multiple objects and maintaining their identities over time.
Recent query-propagation trackers carry target-specific representations across frames and update them with current-frame image features, yet identity association remains difficult when visually similar targets overlap or cross.
During such interactions, current-frame query updates may contain components directed toward nearby active tracks, which may affect identity association in later frames.
To investigate this issue, we analyze track-query updates preceding identity switches (IDSWs), where the same target is assigned a different track ID.
Our analysis shows that
(i) propagated track-query representations change more abruptly before IDSWs,
(ii) current-frame updates have larger positive projections along directions toward other active tracks, and
(iii) bounding-box intersection over union (IoU) retrieves the counterpart track involved in a subsequent IDSW with high recall when that track remains active before the switch.
Motivated by these findings, we propose \textbf{T}est-time Track-\textbf{Q}uery \textbf{C}alibration (TQC), a training-free and gradient-free method for query-propagation trackers.
TQC calibrates the tracker output before propagation using a bounded correction estimated from update components directed toward nearby active tracks.
It requires only track-query representations and bounding boxes available during inference, without retraining or architectural modification.
Across nine DanceTrack and SportsMOT test settings, TQC improves HOTA and AssA in all settings and IDF1 in eight, with an average HOTA gain of $+0.38$ and limited inference overhead.
Controlled variants on six validation settings further show that TQC's correction direction yields more consistent HOTA improvements than alternative directions with matched correction norms.

\end{abstract}
\section{Introduction}

Multi-Object Tracking (MOT) is a fundamental video understanding task for estimating object trajectories in dynamic scenes, with applications in autonomous driving~\cite{yu2020bdd100k}, visual surveillance~\cite{tang2019cityflow,zhu2018vision}, and sports analytics~\cite{scott2024teamtrack,yang2026soccermaster,cioppa2022soccernet}.
It requires localizing multiple objects in each frame and maintaining their identities over time, allowing dynamic scenes to be analyzed at the level of individual objects.
Reliable identity association is crucial because identity errors corrupt trajectories and degrade downstream video analysis.

\begin{figure}[t]
\centering
\includegraphics[width=0.95\linewidth]{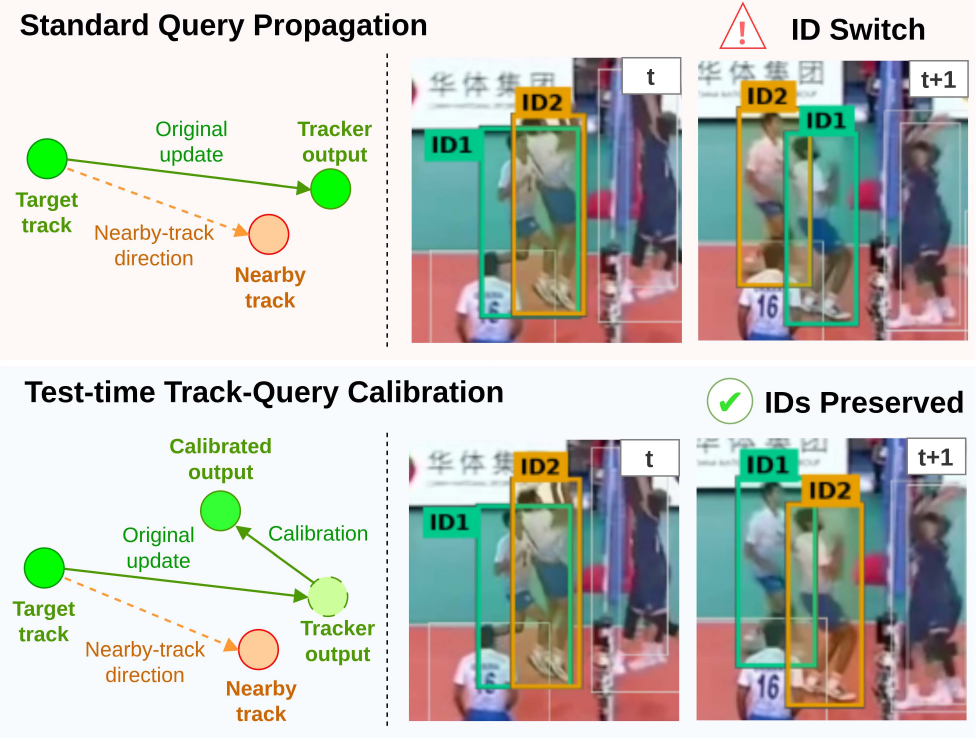}
\caption{
\textbf{Standard query propagation and Test-time Track-Query Calibration (TQC).}
Left: A current-frame query update may contain a component directed toward a nearby active-track representation.
TQC calibrates the tracker output before propagation using a bounded correction estimated from update components directed toward nearby active tracks.
Right: In this example, standard query propagation leads to an identity switch at frame $t+1$, whereas TQC preserves the identities.
}
\label{fig:fig1}
\end{figure}

A growing line of MOT methods supports identity association by propagating track queries or track states across frames~\cite{meinhardt2022trackformer,zeng2022motr,zhang2023motrv2}.
At each frame, these representations are updated with current-frame image features to adapt to changes in target motion and appearance while remaining associated with the corresponding targets.
By carrying target-specific information over time, this formulation enables trackers to reuse that information under temporary occlusion, reappearance, and visual ambiguity.
Nevertheless, preserving a stable correspondence between each propagated representation and its target remains challenging when visually similar targets become spatially close, overlap, or cross, even if each target is correctly detected.

One possible source of this difficulty lies in the current-frame query update.
Ideally, the update should adapt the propagated representation to changes in the corresponding target's motion and appearance while keeping it associated with the same target.
However, during close interactions, the image features used for the update may also contain cues from nearby active tracks.
The resulting update may therefore contain components directed toward nearby active-track representations.
Because the updated representation is propagated to subsequent frames, such components may affect identity association in later frames.

To investigate this issue, we analyze track-query updates before identity switches (IDSWs), where the same target is assigned a different track ID.
Our analysis first shows that propagated track-query representations change more abruptly before IDSWs.
However, the magnitude of this change alone does not reveal whether it reflects target-specific adaptation or contains a component directed toward another active track.
We therefore examine the alignment between the current-frame query update and directions toward other active tracks in the query representation space.
Updates before IDSWs have larger positive projections along these directions.
We further show that bounding-box intersection over union (IoU) retrieves the counterpart track involved in a subsequent IDSW with high recall when that track remains active before the switch.
Updates before IDSWs also have larger positive projections along directions toward these IoU-based nearby candidates.
Together, these findings suggest that abrupt track-query changes before IDSWs partly reflect update components directed toward nearby active tracks rather than only target-specific adaptation.

Motivated by these findings, we propose \textbf{T}est-time Track-\textbf{Q}uery \textbf{C}alibration (TQC), a training-free and gradient-free method for query-propagation trackers.
As illustrated in Fig.~\ref{fig:fig1}, TQC calibrates the tracker output before propagation using a bounded correction estimated from update components directed toward nearby active tracks.
Specifically, it retrieves nearby active tracks using bounding-box IoU and estimates the update components directed toward these tracks.
TQC uses only track-query representations and bounding boxes already available during inference, allowing it to be integrated into existing query-propagation trackers without retraining, architectural modifications, or external memory.

Our contributions are summarized as follows:
\begin{itemize}

\item We analyze track-query updates before IDSWs and show that propagated track-query representations change more abruptly, while current-frame updates have larger positive projections toward other active tracks.
We further show that IoU-based candidates retrieve the counterpart track with high recall when it remains active before the switch.

\item We propose Test-time Track-Query Calibration (TQC), a training-free and gradient-free method that estimates update components directed toward nearby active tracks and subtracts a bounded correction from the tracker output before propagation.

\item We demonstrate that TQC improves HOTA and AssA in all nine DanceTrack and SportsMOT test settings, with average gains of $+0.38$ HOTA, $+0.48$ AssA, and $+0.49$ IDF1.
TQC also improves HOTA in all nine settings on association-hard validation subsets under two selection criteria.
Controlled variants support the importance of the correction direction, while TQC adds only $0.89$ to $1.37$ ms of inference latency per frame.

\end{itemize}
\section{Related Work}
\label{sec:related}

\subsection{Multi-Object Tracking}

Multi-Object Tracking (MOT) has traditionally been addressed through tracking-by-detection.
In this paradigm, frame-level detections are first obtained and then associated across frames using motion models, appearance embeddings, IoU, confidence scores, or their combinations~\cite{bewley2016simple,wojke2017simple,zhang2021fairmot,aharon2022bot,zhang2022bytetrack,cao2023observation}.

A prominent line of end-to-end MOT methods instead maintains target-specific track queries or track states and propagates them across frames.
TrackFormer~\cite{meinhardt2022trackformer} formulates MOT as frame-to-frame set prediction by propagating track predictions through a video sequence, while MOTR~\cite{zeng2022motr} transfers and updates track queries across frames.
By carrying target-specific information across frames, this formulation helps maintain identities under temporary occlusion, reappearance, and visual ambiguity.
Because each updated representation is propagated to subsequent frames, errors in the current-frame query update can affect identity association in later frames.

Recent end-to-end MOT methods address identity association in different ways.
Some enrich propagated representations through long-term memory, multi-tracklet temporal modeling, or multiple queries per target~\cite{gao2023memotr,segu2025samba,zeng2025tgformer}.
Others introduce new training-assignment strategies or learnable association mechanisms~\cite{luo2025co,wang2025motr}, or improve target representations through discriminative object embeddings and motion-aware target representation learning~\cite{shao2026detection,huang2026motra}.

More directly related to track-query updates, Dual-Path Temporal Decoder~\cite{kim2025dual} separates appearance adaptation from identity preservation and introduces confidence-guided update suppression for unreliable predictions.
MATR~\cite{yang2025motion} uses inter-frame motion prediction to update track-query features and positional embeddings before decoding, aiming to reduce query collisions.
FADE~\cite{bouzidi2026out} studies adversarial query-state manipulation in propagation-based trackers to induce identity switches and track termination.

TQC operates on the current-frame query update of an existing query-propagation tracker at inference time.
It retrieves nearby active tracks using bounding-box IoU, estimates update components directed toward these tracks, and subtracts a bounded correction before propagation without additional training or architectural modification.

\subsection{Test-Time Adaptation and Calibration}

Test-Time Adaptation (TTA) aims to address distribution shifts by adapting source-trained models to unlabeled test data during inference.
Representative methods include Tent~\cite{wang2021tent}, which estimates normalization statistics and optimizes normalization-layer affine parameters through entropy minimization, and CoTTA~\cite{wang2022continual}, which mitigates error accumulation and catastrophic forgetting in continual adaptation.
TTA has been studied in object detection~\cite{ruan2024fully}, image segmentation~\cite{chen2024each}, and vision-language models~\cite{han2025dota}.

Test-time methods have also been explored in MOT~\cite{segu2023darth,guo2026dual}.
DARTH~\cite{segu2023darth} performs holistic test-time adaptation for MOT by adapting detection and appearance representations under domain shift using detection consistency and patch contrastive losses.
It is a gradient-based adaptation method mainly designed for appearance-based tracking-by-detection pipelines~\cite{pang2021quasi}.
TCEI~\cite{guo2026dual} performs test-time calibration for MOT using transient memory for recently observed objects and accumulated experience from prior test videos to recalibrate identity predictions.
It is built on MOTIP~\cite{gao2025multiple}, which formulates MOT as in-context ID prediction rather than query propagation.

These methods differ from TQC in their target problem and correction mechanism.
DARTH targets detection and appearance representations under domain shift in tracking-by-detection pipelines, while TCEI targets ID-prediction maps in the MOTIP-based ID-prediction framework and recalibrates them using additional memory structures.
In contrast, TQC does not assume a distribution shift and instead targets the current-frame query update in query-propagation trackers before the updated representation is propagated.
It uses only query representations and bounding boxes already available during inference, without model-parameter updates, gradients, or additional memory structures.
\begin{figure*}[t]
\centering
\setlength{\tabcolsep}{2pt}
\begin{tabular}{ccc}
\includegraphics[width=0.315\textwidth]{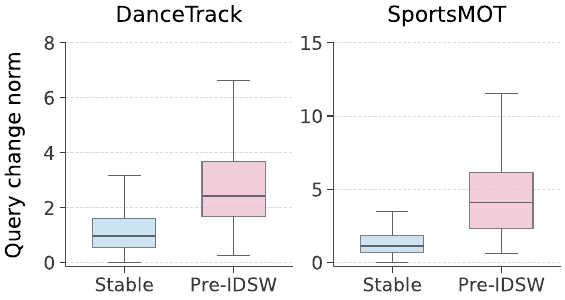} &
\includegraphics[width=0.315\textwidth]{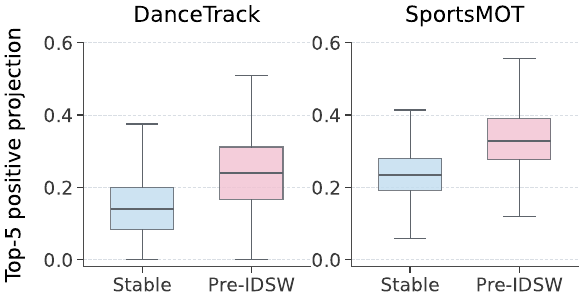} &
\includegraphics[width=0.315\textwidth]{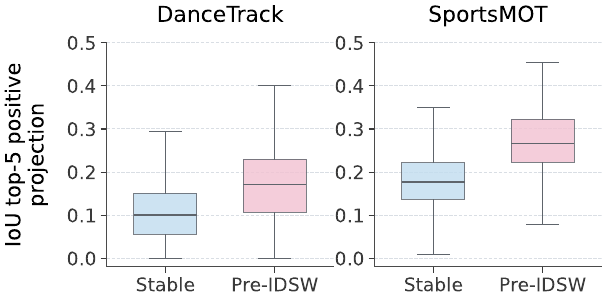} \\
\small (a) & \small (b) & \small (c)
\end{tabular}
\caption{
\textbf{Distributions of query-update statistics before IDSWs.}
(a) Norm of the frame-to-frame change in propagated query representations.
(b) Average of the five largest normalized positive projections over all other active tracks.
(c) Average normalized positive projection over IoU top-5 nearby active tracks.
}
\label{fig:query_update_distributions}
\end{figure*}

\section{Analysis of Track-Query Updates Before Identity Switches}
\label{sec:analysis}

We analyze track-query updates preceding identity switches (IDSWs) in query-propagation trackers.
An IDSW occurs when the tracker assigns a different track ID to the same ground-truth target.
We focus on IDSWs because they directly reflect failures in identity association.
Unless otherwise stated, we conduct the analysis using the DAB-DETR~\cite{liu2022dab} variant of MeMOTR~\cite{gao2023memotr} on the DanceTrack and SportsMOT validation sets.
We additionally verify the findings with SambaMOTR~\cite{segu2025samba}, with detailed results provided in the supplementary material.

\subsection{Propagated Track Queries Change More Abruptly Before IDSWs}

We first measure how propagated track-query representations change across consecutive frames.
For track $i$, let $q_i^t$ denote the propagated track-query representation at frame $t$, before incorporating current-frame features.
We define the change as:
\begin{equation}
\Delta q_i^t = q_i^{t+1} - q_i^t .
\label{eq:delta_q}
\end{equation}

For analysis, we consider true-positive track states, focusing on identity-association behavior rather than missed detections or false positives.
We exclude states for which an IDSW occurs at frame $t$, so that pre-IDSW states precede rather than coincide with the switch.
Among the remaining states, we label a state at frame $t$ as pre-IDSW if an IDSW occurs for the same ground-truth trajectory within the next $H$ frames, and as stable otherwise.
We use $H=3$ throughout the analysis.
These labels are used only for analysis and are not used by TQC.

Fig.~\ref{fig:query_update_distributions}(a) compares the distributions of $\lVert \Delta q_i^t \rVert_2$ for stable and pre-IDSW states.
Pre-IDSW states exhibit larger query changes than stable states on both datasets.
On DanceTrack, the mean norm increases from $1.244$ for stable states to $2.884$ for pre-IDSW states, with an AUROC of $0.835$.
On SportsMOT, the corresponding mean norms are $1.466$ and $4.645$, with an AUROC of $0.892$.
These results show that propagated track-query representations change more abruptly before IDSWs.

\subsection{Query Updates Show Stronger Positive Projections Toward Other Active Tracks Before IDSWs}

A large frame-to-frame query change is not necessarily harmful.
Such changes may be required to follow abrupt motion, pose variation, appearance change, or recovery from occlusion.
The magnitude of $\Delta q_i^t$ alone therefore does not reveal whether the change reflects target-specific adaptation or contains a component directed toward another active track.
To examine whether such components are stronger before IDSWs, we measure the alignment between current-frame query updates and directions toward other active tracks.

Let $o_i^t$ denote the tracker output representation after incorporating current-frame image features.
We define the current-frame query update as:
\begin{equation}
d_i^t = o_i^t - q_i^t .
\label{eq:update_direction}
\end{equation}
For another active track $j$, we define the direction from track $i$ to track $j$ in the query representation space as:
\begin{equation}
u_{ij}^t =
\frac{q_j^t - q_i^t}
{\lVert q_j^t - q_i^t\rVert_2} .
\label{eq:candidate_direction}
\end{equation}

We quantify this alignment using the normalized positive projection:
\begin{equation}
\bar{p}_{ij}^t =
\frac{\left[(d_i^t)^\top u_{ij}^t\right]_+}
{\lVert d_i^t\rVert_2} ,
\label{eq:normalized_positive_projection}
\end{equation}
where $[\cdot]_+ = \max(\cdot,0)$.
This score measures how strongly the update of track $i$ is aligned with the direction toward track $j$, independently of the update magnitude.
Negative projections are ignored because they correspond to updates moving away from track $j$.
For each track $i$, we compute $\bar{p}_{ij}^t$ over all other active tracks and average the five largest values.

Fig.~\ref{fig:query_update_distributions}(b) shows that the average of the top-5 positive projections is larger for pre-IDSW states than for stable states.
On DanceTrack, the mean score increases from $0.147$ to $0.240$, with an AUROC of $0.759$.
On SportsMOT, the corresponding means are $0.239$ and $0.337$, with an AUROC of $0.827$.
Thus, current-frame query updates have larger positive projections along directions toward other active tracks before IDSWs.
The supplementary material further shows that this separation persists after covariance whitening of the query representations.

\begin{figure}[t]
\centering
\includegraphics[width=0.9\linewidth]{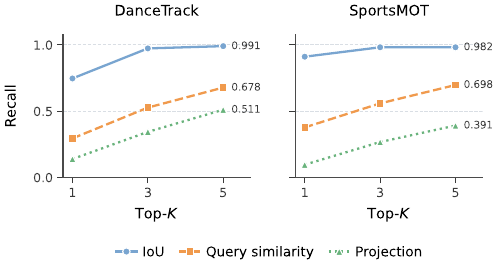}
\caption{
\textbf{Candidate retrieval recall for IDSW counterpart tracks.}
For each evaluated pre-IDSW state in which the counterpart track is active, other active tracks are ranked by bounding-box IoU, query cosine similarity, or the normalized positive projection in Eq.~\ref{eq:normalized_positive_projection}.
}
\label{fig:candidate_retrieval_recall}
\end{figure}

\begin{figure*}[t]
\centering
\includegraphics[width=\linewidth]{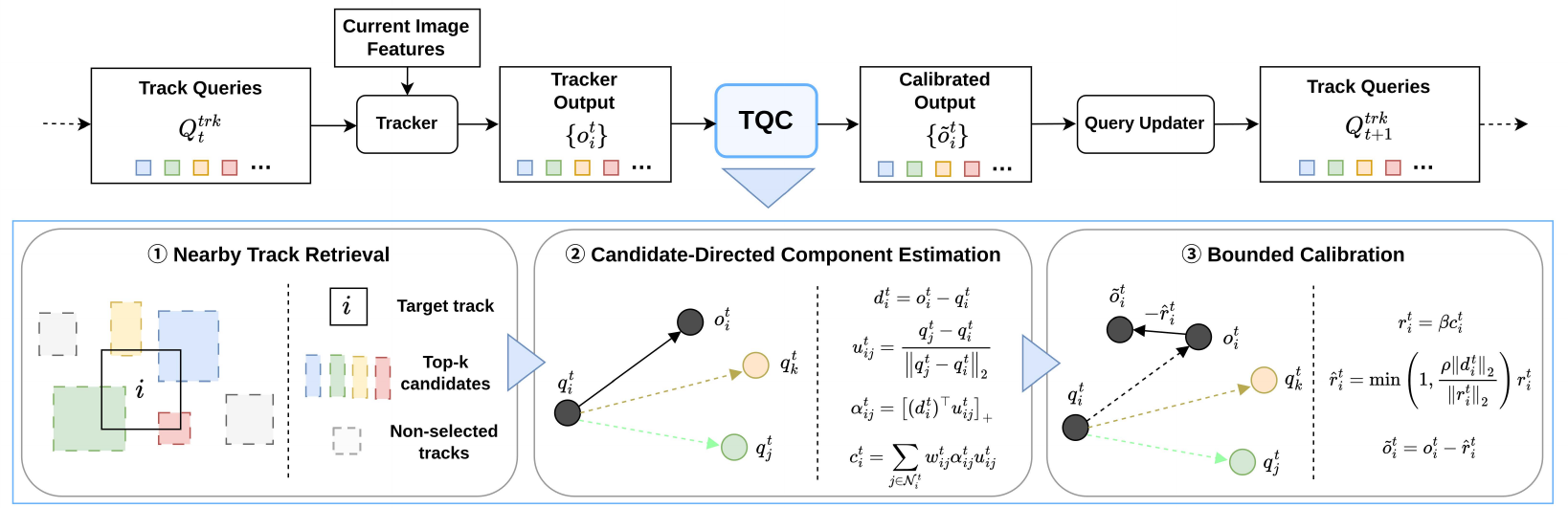}
\caption{
\textbf{Overview of TQC.}
The upper flow shows the standard query-propagation process, in which track-query representations are updated with current-frame image features and propagated to the next frame.
TQC is inserted after the tracker forward pass and calibrates the tracker output representations before subsequent propagation.
It retrieves nearby active tracks using bounding-box IoU, estimates update components directed toward these tracks, and subtracts a bounded correction.
}
\label{fig:tqc_overview}
\end{figure*}

\subsection{IoU Retrieves IDSW Counterpart Tracks}

We next examine how the counterpart track involved in a subsequent IDSW can be retrieved from current-frame information.
Spatial proximity provides a natural cue for this retrieval, while representation similarity and update-direction alignment provide alternative cues.
We therefore compare bounding-box IoU, query cosine similarity, and the normalized positive projection.

For an IDSW event, we refer to the track assigned to the same ground-truth trajectory after the switch as the counterpart track.
Because retrieval can only be evaluated when the counterpart track is active in the current frame, we consider pre-IDSW states in which it is active.
For each evaluated pre-IDSW state, the other active tracks in the same frame are ranked according to each criterion.
Fig.~\ref{fig:candidate_retrieval_recall} reports top-$K$ recall for retrieving the counterpart track.
IoU achieves the highest top-$K$ recall among the three criteria.
In particular, IoU top-5 candidates include the counterpart track in $99.1\%$ of evaluated states on DanceTrack and $98.2\%$ on SportsMOT.
Based on this result, TQC uses bounding-box IoU to identify nearby active tracks.

We then compute the average normalized positive projection over the IoU top-5 candidates.
As shown in Fig.~\ref{fig:query_update_distributions}(c), this score is larger for pre-IDSW states than for stable states.
On DanceTrack, the mean score increases from $0.109$ to $0.171$, with an AUROC of $0.713$.
On SportsMOT, the corresponding means are $0.182$ and $0.272$, with an AUROC of $0.829$.
Together, the high retrieval recall and larger positive projections before IDSWs motivate the IoU-based candidate selection in TQC.
\section{Test-Time Track-Query Calibration}
\label{sec:method}

We propose \textbf{T}est-time Track-\textbf{Q}uery \textbf{C}alibration (TQC), a training-free and gradient-free method for query-propagation trackers.

\subsection{Overview}
\label{sec:method_overview}

Our analysis in Sec.~\ref{sec:analysis} shows that current-frame query updates have larger normalized positive projections toward other active tracks before IDSWs.
Bounding-box IoU also retrieves the counterpart track involved in a subsequent IDSW with high recall when it is active.
Motivated by these findings, TQC selects nearby active tracks using IoU and subtracts a bounded correction estimated from update components directed toward these tracks before propagation, as illustrated in Fig.~\ref{fig:tqc_overview}.

For track $i$, let $q_i^t$ denote the propagated track-query representation before incorporating current-frame image features, and let $o_i^t$ denote the corresponding tracker output representation.
Their difference $d_i^t = o_i^t - q_i^t$ represents the current-frame query update.

TQC uses only track-query representations, tracker output representations, and current-frame bounding boxes already available during inference.
It requires no additional training or gradient computation and does not modify the tracker architecture.

\subsection{Candidate-Directed Component Estimation}

Let $\mathcal{A}^t$ denote the set of active tracks whose query representations and bounding boxes are available at frame $t$.
For each continuing track $i$, TQC retrieves nearby active tracks using bounding-box IoU.
Newly initialized tracks are not themselves calibrated, but they can be included as candidates for other tracks.
Specifically, TQC selects up to $K$ active tracks with the largest positive IoU:
\begin{equation}
\mathcal{N}_i^t =
\operatorname{TopK}_{j \in \mathcal{A}^t \setminus \{i\},\ 
\operatorname{IoU}(b_i^t,b_j^t)>0}
\operatorname{IoU}(b_i^t,b_j^t).
\label{eq:candidate_retrieval}
\end{equation}
Here, $b_i^t$ and $b_j^t$ denote the current-frame bounding boxes produced by the standard tracker forward pass.
If no candidate with positive IoU exists, TQC leaves the output representation unchanged.
We use $K=5$ by default, following the candidate-retrieval analysis in Sec.~\ref{sec:analysis}.

For each candidate $j \in \mathcal{N}_i^t$, TQC defines the unit direction from track $i$ toward track $j$ in the query representation space:
\begin{equation}
u_{ij}^t =
\frac{q_j^t - q_i^t}
{\lVert q_j^t - q_i^t\rVert_2}.
\label{eq:method_candidate_direction}
\end{equation}
The positive projection of the query update onto this direction is:
\begin{equation}
\alpha_{ij}^t =
\left[(d_i^t)^\top u_{ij}^t\right]_+ ,
\label{eq:positive_projection_magnitude}
\end{equation}
where $[\cdot]_+ = \max(\cdot,0)$.
The value $\alpha_{ij}^t$ measures the positive projection of the update of track $i$ along the direction toward candidate track $j$.
Updates moving away from the candidate do not contribute to the correction.

TQC assigns larger weights to candidates with stronger spatial overlap:
\begin{equation}
w_{ij}^t =
\frac{
\operatorname{IoU}(b_i^t,b_j^t)
}{
\sum_{k \in \mathcal{N}_i^t}
\operatorname{IoU}(b_i^t,b_k^t)
}.
\label{eq:iou_weight}
\end{equation}
The weighted components are then aggregated as:
\begin{equation}
c_i^t =
\sum_{j \in \mathcal{N}_i^t}
w_{ij}^t \alpha_{ij}^t u_{ij}^t .
\label{eq:iou_weighted_correction}
\end{equation}
By construction, $c_i^t$ has a nonnegative inner product with the current-frame query update:
\begin{equation}
(d_i^t)^\top c_i^t
=
\sum_{j \in \mathcal{N}_i^t}
w_{ij}^t
(\alpha_{ij}^t)^2
\ge 0.
\label{eq:correction_alignment}
\end{equation}
Because the candidate directions are not necessarily orthogonal, this property does not guarantee that the positive projection onto every candidate direction decreases after calibration.

\subsection{Bounded Calibration}
\label{sec:bounded_calibration}

TQC constructs a correction from $c_i^t$ and subtracts it from the tracker output representation before propagation.
Because $c_i^t$ aggregates components along multiple candidate directions, directly applying a large correction may also remove useful parts of the current-frame query update.
TQC therefore controls the correction using a strength parameter $\beta$ and bounds its magnitude relative to the original update:
\begin{equation}
r_i^t = \beta c_i^t,
\qquad
\hat{r}_i^t =
\min
\left(
1,
\frac{\rho \lVert d_i^t\rVert_2}
{\lVert r_i^t\rVert_2}
\right)
r_i^t .
\label{eq:bounded_correction}
\end{equation}
Here, $\beta>0$ controls the correction strength and $0<\rho\leq1$ specifies the maximum correction magnitude relative to the original update.
If $r_i^t=0$, we define $\hat{r}_i^t=0$.
The resulting correction therefore satisfies:
\begin{equation}
\lVert \hat{r}_i^t \rVert_2
\leq
\rho \lVert d_i^t \rVert_2 .
\label{eq:correction_bound}
\end{equation}

TQC calibrates the tracker output representation as:
\begin{equation}
\tilde{o}_i^t =
o_i^t - \hat{r}_i^t .
\label{eq:calibrated_query}
\end{equation}
The calibrated representation $\tilde{o}_i^t$ is used in place of $o_i^t$ for subsequent propagation.
\section{Experiments}
\label{sec:experiments}

We evaluate TQC across multiple query-propagation trackers on DanceTrack and SportsMOT.
We first report test-set performance and results on association-hard validation sequences, followed by analyses of the correction direction, its effect on query updates, qualitative examples, and inference efficiency.
Additional results and analyses, including further dataset evaluations, hyperparameter sensitivity, and detailed identity-association statistics, are provided in the supplementary material.

\subsection{Experimental Setup}

\noindent\textbf{Datasets.}
We evaluate TQC on DanceTrack~\cite{sun2022dancetrack} and SportsMOT~\cite{cui2023sportsmot}, which contain visually similar targets, complex motion, and frequent target interactions.

\noindent\textbf{Trackers and implementation details.}
We evaluate six query-propagation tracker configurations, MOTR~\cite{zeng2022motr}, MeMOTR-Deformable~\cite{gao2023memotr}, CO-MOT~\cite{luo2025co}, MeMOTR-DAB, SambaMOTR-DAB~\cite{segu2025samba}, and DualMOT~\cite{kim2025dual}.
Here, Deformable and DAB denote the Deformable DETR~\cite{zhu2021deformable} and DAB-DETR~\cite{liu2022dab} variants, respectively.
MOTR, CO-MOT, and DualMOT are evaluated on DanceTrack, while the remaining three configurations are evaluated on both datasets, resulting in nine tracker and dataset settings.
Unless otherwise specified, TQC uses $K=5$ candidate tracks selected from active tracks with positive bounding-box IoU.
The correction parameters $\beta$ and $\rho$ are selected using validation HOTA for each tracker and dataset setting and then fixed for test-set evaluation.
The selected values are provided in the supplementary material.

\noindent\textbf{Evaluation metrics.}
We report HOTA~\cite{luiten2021hota}, DetA~\cite{luiten2021hota}, AssA~\cite{luiten2021hota}, MOTA~\cite{bernardin2008evaluating}, and IDF1~\cite{ristani2016performance}.
We mainly discuss HOTA, AssA, and IDF1, where HOTA summarizes overall tracking quality, while AssA and IDF1 reflect identity association quality.

\begin{table}[t]
\centering
\caption{
\textbf{Test results on DanceTrack.}
$^\dagger$ DualMOT is evaluated using the publicly released checkpoint without confidence-guided update suppression.
Reproduction details are provided in the supplementary material.
}
\renewcommand{\arraystretch}{0.90}
\setlength{\tabcolsep}{3.0pt}
\begin{tabular}{@{}lccccc@{}}
\toprule
Method
& HOTA
& DetA
& AssA
& MOTA
& IDF1 \\
\midrule
\multicolumn{6}{@{}l}{\textit{Deformable:}} \\
\addlinespace[0.2em]
MOTR
& 54.21 & 73.49 & 40.19 & 79.68 & 51.49 \\
\quad + TQC
& \textbf{54.92} & \textbf{73.79} & \textbf{41.08}
& \textbf{80.07} & \textbf{52.30} \\
\addlinespace[0.25em]
MeMOTR
& 63.24 & 76.95 & 52.13 & 85.45 & 65.43 \\
\quad + TQC
& \textbf{64.15} & \textbf{77.55} & \textbf{53.24}
& \textbf{86.12} & \textbf{66.30} \\
\addlinespace[0.25em]
CO-MOT
& 69.96 & \textbf{82.07} & 59.78 & 90.99 & 72.25 \\
\quad + TQC
& \textbf{70.16} & 82.01 & \textbf{60.17}
& \textbf{91.33} & \textbf{72.96} \\
\midrule
\multicolumn{6}{@{}l}{\textit{DAB:}} \\
\addlinespace[0.2em]
MeMOTR
& 68.51 & 80.52 & 58.47 & 89.88 & 71.20 \\
\quad + TQC
& \textbf{68.92} & \textbf{80.85} & \textbf{58.91}
& \textbf{90.33} & \textbf{72.32} \\
\addlinespace[0.25em]
SambaMOTR
& 68.76 & 79.28 & 59.84 & 89.06 & 73.64 \\
\quad + TQC
& \textbf{69.08} & \textbf{79.50} & \textbf{60.21}
& \textbf{89.10} & \textbf{73.89} \\
\midrule
DualMOT$^\dagger$
& 72.88 & 82.52 & 64.49 & 91.51 & \textbf{78.27} \\
\quad + TQC
& \textbf{73.09} & \textbf{82.65} & \textbf{64.74}
& \textbf{91.66} & 78.02 \\
\bottomrule
\end{tabular}
\label{tab:dancetrack_test_tqc}
\end{table}

\begin{table}[t]
\centering
\caption{
\textbf{Test results on SportsMOT.}
}
\renewcommand{\arraystretch}{0.90}
\setlength{\tabcolsep}{3.0pt}
\begin{tabular}{@{}lccccc@{}}
\toprule
Method
& HOTA
& DetA
& AssA
& MOTA
& IDF1 \\
\midrule
\multicolumn{6}{@{}l}{\textit{Deformable:}} \\
\addlinespace[0.2em]
MeMOTR
& 68.78 & 82.01 & 57.77 & 90.21 & 69.89 \\
\quad + TQC
& \textbf{69.19} & \textbf{82.15} & \textbf{58.36}
& \textbf{90.40} & \textbf{70.46} \\
\midrule
\multicolumn{6}{@{}l}{\textit{DAB:}} \\
\addlinespace[0.2em]
MeMOTR
& 70.00 & 83.04 & 59.09 & 91.46 & 71.38 \\
\quad + TQC
& \textbf{70.14} & \textbf{83.14} & \textbf{59.26}
& \textbf{91.64} & \textbf{71.53} \\
\addlinespace[0.25em]
SambaMOTR
& 72.14 & 83.39 & 62.49 & 91.90 & 74.79 \\
\quad + TQC
& \textbf{72.22} & \textbf{83.47} & \textbf{62.56}
& \textbf{91.99} & \textbf{74.96} \\
\bottomrule
\end{tabular}
\label{tab:sportsmot_test_tqc}
\end{table}

\begin{table}[t]
\centering
\caption{
\textbf{Performance on association-hard validation sequences.}
For each tracker and dataset setting, we independently select ten sequences under each of two baseline criteria.
Values are mean TQC improvements across tracker settings, computed from sequence-averaged metrics.
Pos. denotes the number of settings with positive $\Delta$HOTA.
}
\small
\setlength{\tabcolsep}{3.0pt}
\begin{tabular}{@{}lrrrc@{}}
\toprule
Dataset
& $\Delta$HOTA
& $\Delta$AssA
& $\Delta$IDF1
& Pos. \\
\midrule
\multicolumn{5}{@{}l}{
\textit{Largest baseline $\mathrm{DetA}-\mathrm{AssA}$}} \\
DanceTrack
& +1.53 & +2.03 & +1.88 & 6/6 \\
SportsMOT
& +1.07 & +1.72 & +1.27 & 3/3 \\
Overall
& \textbf{+1.37} & \textbf{+1.93}
& \textbf{+1.68} & \textbf{9/9} \\
\midrule
\multicolumn{5}{@{}l}{
\textit{Lowest baseline AssA}} \\
DanceTrack
& +1.40 & +1.71 & +1.74 & 6/6 \\
SportsMOT
& +1.11 & +1.71 & +1.43 & 3/3 \\
Overall
& \textbf{+1.30} & \textbf{+1.71}
& \textbf{+1.64} & \textbf{9/9} \\
\bottomrule
\end{tabular}
\label{tab:association_hard_summary}
\end{table}

\begin{table}[t]
\centering
\caption{
\textbf{Controlled variants on six validation settings.}
Values are average changes from the corresponding baselines.
Pos. denotes the number of settings with positive $\Delta$HOTA.
Direction variants preserve the TQC correction norm, and random directions are averaged over eight seeds.
Per-setting results are provided in the supplementary material.
}
\footnotesize
\renewcommand{\arraystretch}{1.05}
\setlength{\tabcolsep}{2.5pt}
\begin{tabular}{@{}lrrrc@{}}
\toprule
Variant
& $\Delta$HOTA
& $\Delta$AssA
& $\Delta$IDF1
& Pos. \\
\midrule
\textbf{Full TQC}
& \textbf{+0.71} & \textbf{+1.06} & \textbf{+0.88} & \textbf{6/6} \\
\midrule
\multicolumn{5}{@{}l}{\textit{Correction direction}} \\
Reversed correction
& -0.91 & -1.28 & -1.36 & 0/6 \\
Random directions
& -0.04 & -0.05 & -0.07 & 3/6 \\
Orthogonal directions
& -0.06 & -0.09 & +0.02 & 2/6 \\
\shortstack[l]{Non-overlapping-track\\directions}
& +0.10 & +0.15 & +0.01 & 4/6 \\
\midrule
\multicolumn{5}{@{}l}{\textit{Calibration strategy}} \\
Norm-preserving correction
& +0.30 & +0.35 & +0.33 & 4/6 \\
Update damping
& +0.29 & +0.48 & +0.39 & 5/6 \\
Uniform damping ($\gamma=0.95$)
& +0.30 & +0.45 & +0.42 & 5/6 \\
Uniform damping ($\gamma=0.75$)
& +0.11 & +0.10 & +0.19 & 5/6 \\
Uniform damping ($\gamma=0.50$)
& -1.33 & -1.90 & -1.70 & 0/6 \\
\midrule
\multicolumn{5}{@{}l}{\textit{Candidate weighting}} \\
w/o IoU weights
& +0.15 & +0.25 & +0.21 & 4/6 \\
Projection weights
& +0.13 & +0.24 & +0.04 & 4/6 \\
\midrule
\multicolumn{5}{@{}l}{\textit{Correction bound}} \\
w/o bound
& -2.91 & -3.70 & -3.33 & 2/6 \\
\midrule
\multicolumn{5}{@{}l}{\textit{Number of candidates}} \\
$K=3$
& +0.60 & +0.89 & +0.70 & 5/6 \\
$K=10$
& +0.69 & +1.03 & +0.77 & 6/6 \\
\bottomrule
\end{tabular}
\label{tab:controls_six_settings}
\end{table}

\begin{table}[t]
\centering
\caption{
\textbf{Effect of TQC on query updates.}
Projection and update-norm retention are state-wise ratios between the TQC branch and the paired branch with the current-frame correction disabled.
}
\small
\renewcommand{\arraystretch}{0.90}
\setlength{\tabcolsep}{3.5pt}
\begin{tabular}{@{}lcc@{}}
\toprule
& \shortstack{Tracker output\\after TQC}
& \shortstack{Track query used\\for propagation} \\
\midrule
Projection decreased
& $98.38\%$
& $97.21\%$ \\
Median projection retention
& $0.317$
& $0.235$ \\
Median update-norm retention
& $0.980$
& $0.792$ \\
\bottomrule
\end{tabular}
\label{tab:tqc_query_effect}
\end{table}

\subsection{Main Results}

Tables~\ref{tab:dancetrack_test_tqc} and~\ref{tab:sportsmot_test_tqc} report test-set results on DanceTrack and SportsMOT.
Across all nine tracker and dataset settings, TQC improves HOTA and AssA in every setting, with average gains of $+0.38$ HOTA, $+0.48$ AssA, and $+0.49$ IDF1.
The HOTA gains range from $+0.20$ to $+0.91$ on DanceTrack and from $+0.08$ to $+0.41$ on SportsMOT.
IDF1 improves in eight of the nine settings, with the only decrease observed for DualMOT$^\dagger$.
TQC also reduces IDSWs in eight of the nine settings, with detailed counts provided in the supplementary material.
We further evaluate TQC on association-hard validation sequences, selecting the ten sequences with the largest baseline $\mathrm{DetA}-\mathrm{AssA}$ or the lowest baseline AssA for each tracker and dataset setting.
As shown in Table~\ref{tab:association_hard_summary}, TQC improves HOTA in all nine settings under both criteria, with average gains of $+1.37$ and $+1.30$, respectively.
Under the largest-gap criterion, the average HOTA gains are $+1.53$ on DanceTrack and $+1.07$ on SportsMOT, including $+1.48$ for CO-MOT and $+1.11$ for DualMOT on DanceTrack.

\begin{figure*}[t]
\centering
\includegraphics[width=\linewidth]{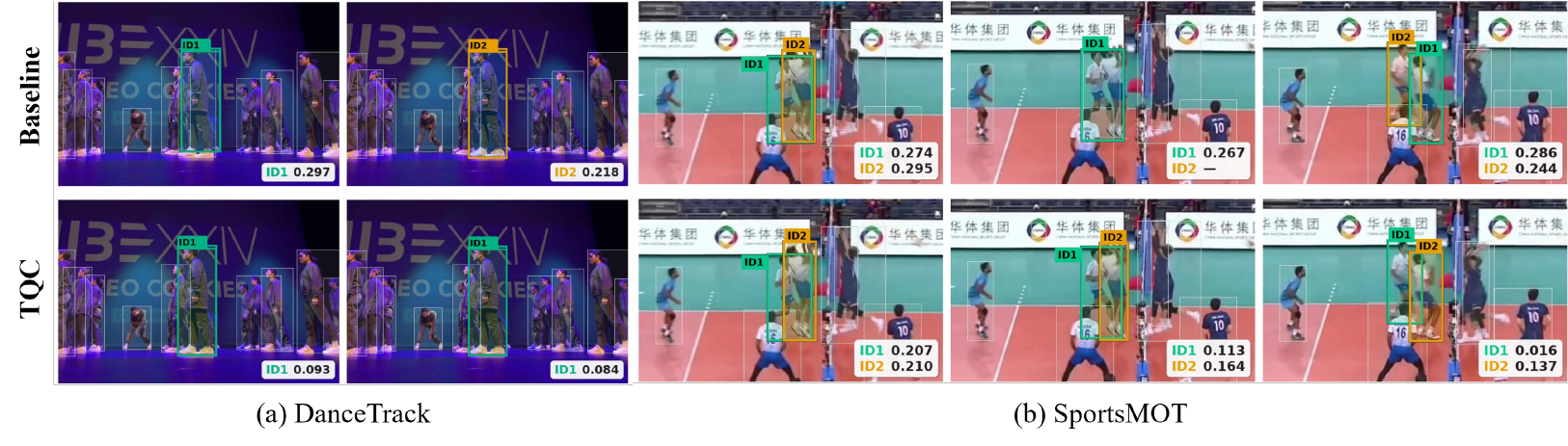}
\caption{
\textbf{Qualitative examples of TQC.}
Examples from DanceTrack and SportsMOT are shown using MeMOTR-DAB.
Values denote the mean normalized positive projection over up to five nearby active tracks with positive bounding-box IoU.
}
\label{fig:qualitative_comparison}
\end{figure*}

\subsection{Analysis of TQC}

\noindent\textbf{Correction Direction and Calibration Controls.}
We evaluate controlled variants on six validation settings to examine the importance of the correction direction and the effects of the calibration design.
Direction variants preserve the TQC correction norm to isolate the effect of direction.
As shown in Table~\ref{tab:controls_six_settings}, full TQC improves HOTA in all six settings, with average gains of $+0.71$ HOTA, $+1.06$ AssA, and $+0.88$ IDF1.
In contrast, reversing the correction direction decreases HOTA in all six settings, with an average change of $-0.91$.
Random, orthogonal, and non-overlapping-track directions yield lower average $\Delta$HOTA and fail to improve HOTA consistently across all six settings.
For each of the six settings, all eight random-direction runs yield lower HOTA than full TQC, giving $48/48$ comparisons in total.
These results support the importance of the correction direction for consistent improvements.

We further examine whether changes in the update magnitude alone can explain the improvements.
Norm-preserving correction, update damping, and uniform damping yield smaller average gains than full TQC.
Alternative candidate weighting schemes also yield smaller average gains.
Removing the correction bound substantially degrades performance.
Performance remains relatively stable for $K=3$ and $K=10$ around the default $K=5$.
When $(\beta,\rho)$ are selected on the other dataset's validation set, TQC improves test HOTA in five of six settings, with a mean gain of $+0.15$ compared with $+0.38$ under target-specific selection on the same six settings (see the supplementary material).

\paragraph{Effect on Query Updates.}
We examine how TQC changes query updates using MeMOTR-DAB on all $25$ DanceTrack validation sequences.
For each state, we compare the TQC branch with a paired branch initialized from the same tracker state but without the current-frame correction.
As shown in Table~\ref{tab:tqc_query_effect}, the average positive projection along the selected candidate directions decreases in $98.38\%$ of states immediately after TQC and remains lower in $97.21\%$ of states in the track-query representation used for propagation.
The median projection retention is $0.317$ immediately after TQC and $0.235$ for propagation, while the corresponding update-norm retention is $0.980$ and $0.792$.
Thus, immediately after calibration, TQC substantially reduces the selected positive projections while retaining most of the overall update magnitude.
The decrease is observed in all $25$ validation sequences, with further analyses provided in the supplementary material.

To connect this result to the pre-IDSW analysis in Sec.~\ref{sec:analysis}, we separately examine pre-IDSW states from the baseline trajectories, using the subsequent IDSW counterpart only for evaluation.
Among states in which the counterpart contributes a nonzero correction term, TQC decreases the positive projection toward that counterpart in $752/767$ states ($98.04\%$).
The median projection retention toward the counterpart is $0.378$, while the median update-norm retention is $0.971$.

\paragraph{Qualitative Examples.}
As shown in Fig.~\ref{fig:qualitative_comparison}, TQC yields lower
normalized positive projections toward nearby active tracks in these
examples and preserves the highlighted identities, whereas the baseline changes or swaps identities.
These examples are consistent with the quantitative reductions in
positive projection reported in Table~\ref{tab:tqc_query_effect}.

\paragraph{Efficiency.}
As shown in Table~\ref{tab:runtime_dancetrack_val}, TQC improves test-set HOTA across all six DanceTrack tracker configurations, while adding only $0.89$ to $1.37$ ms per frame on validation, corresponding to a relative slowdown of $0.61\%$ to $1.13\%$.

\begin{table}[t]
\centering
\caption{
\textbf{Inference efficiency and HOTA improvements on DanceTrack.}
Inference speed is measured on the validation set using a DGX Spark, while $\Delta$HOTA denotes the test-set improvement over the corresponding baseline.
Slowdown is computed from unrounded per-frame latencies.
}
\small
\renewcommand{\arraystretch}{0.90}
\setlength{\tabcolsep}{3.0pt}
\begin{tabular}{@{}lcccc@{}}
\toprule
Tracker
& \shortstack{Base\\FPS$\uparrow$}
& \shortstack{+TQC\\FPS$\uparrow$}
& \shortstack{Slowdown\\$\downarrow$}
& \shortstack{Test\\$\Delta$HOTA$\uparrow$} \\
\midrule
MOTR
& 9.44 & 9.35 & 1.02\% & +0.71 \\
MeMOTR-Deformable
& 8.26 & 8.17 & 1.13\% & +0.91 \\
CO-MOT
& 9.33 & 9.26 & 0.83\% & +0.20 \\
MeMOTR-DAB
& 7.29 & 7.22 & 0.95\% & +0.41 \\
SambaMOTR-DAB
& 7.29 & 7.23 & 0.91\% & +0.32 \\
DualMOT
& 6.20 & 6.16 & 0.61\% & +0.20 \\
\bottomrule
\end{tabular}
\label{tab:runtime_dancetrack_val}
\end{table}

\paragraph{Limitations.}
TQC is designed for query-propagation trackers and calibrates continuing tracks using nearby active tracks with positive bounding-box IoU.
It therefore does not directly address all identity-association failures, including those related to track initialization or recovery after track termination.
Although TQC consistently improves HOTA and AssA across the nine test settings, the gains are modest in some settings.
\section{Conclusion}
\label{sec:conclusion}

We presented TQC, a training-free and gradient-free test-time calibration method for query-propagation trackers.
Our analysis showed that propagated track-query representations change more abruptly before IDSWs and that current-frame updates exhibit larger positive projections toward other active tracks.
We further showed that IoU-based candidates retrieve counterpart tracks involved in subsequent IDSWs with high recall when those tracks are already active, supporting the candidate selection used by TQC.
Across nine DanceTrack and SportsMOT test settings, TQC improves HOTA and AssA in all settings and IDF1 in eight, with limited inference overhead.
TQC also improves HOTA in all nine settings on association-hard validation subsets under both baseline selection criteria.
Controlled variants support the importance of the correction direction, while paired analyses show that TQC reduces positive projections toward selected candidates in most evaluated states.

\FloatBarrier
{
    \small
    \bibliographystyle{ieeenat_fullname}
    \bibliography{main}

\begin{thebibliography}{43}
\providecommand{\natexlab}[1]{#1}
\providecommand{\url}[1]{\texttt{#1}}
\expandafter\ifx\csname urlstyle\endcsname\relax
  \providecommand{\doi}[1]{doi: #1}\else
  \providecommand{\doi}{doi: \begingroup \urlstyle{rm}\Url}\fi

\bibitem[Aharon et~al.(2022)Aharon, Orfaig, and Bobrovsky]{aharon2022bot}
Nir Aharon, Roy Orfaig, and Ben-Zion Bobrovsky.
\newblock Bot-sort: Robust associations multi-pedestrian tracking.
\newblock \emph{arXiv preprint arXiv:2206.14651}, 2022.

\bibitem[Bernardin and Stiefelhagen(2008)]{bernardin2008evaluating}
Keni Bernardin and Rainer Stiefelhagen.
\newblock Evaluating multiple object tracking performance: the clear mot metrics.
\newblock \emph{EURASIP Journal on Image and Video Processing}, 2008\penalty0 (1):\penalty0 246309, 2008.

\bibitem[Bewley et~al.(2016)Bewley, Ge, Ott, Ramos, and Upcroft]{bewley2016simple}
Alex Bewley, Zongyuan Ge, Lionel Ott, Fabio Ramos, and Ben Upcroft.
\newblock Simple online and realtime tracking.
\newblock In \emph{2016 IEEE international conference on image processing (ICIP)}, pages 3464--3468. Ieee, 2016.

\bibitem[Bouzidi et~al.(2026)Bouzidi, Liu, Achamyeleh, Iddamsetty, and Al~Faruque]{bouzidi2026out}
Halima Bouzidi, Haoyu Liu, Yonatan Achamyeleh, Praneetsai Iddamsetty, and Mohammad Al~Faruque.
\newblock Out of sight, out of track: Adversarial attacks on propagation-based multi-object trackers via query state manipulation.
\newblock In \emph{Proceedings of the IEEE/CVF Conference on Computer Vision and Pattern Recognition}, pages 13326--13335, 2026.

\bibitem[Cao et~al.(2023)Cao, Pang, Weng, Khirodkar, and Kitani]{cao2023observation}
Jinkun Cao, Jiangmiao Pang, Xinshuo Weng, Rawal Khirodkar, and Kris Kitani.
\newblock Observation-centric sort: Rethinking sort for robust multi-object tracking.
\newblock In \emph{Proceedings of the IEEE/CVF conference on computer vision and pattern recognition}, pages 9686--9696, 2023.

\bibitem[Chen et~al.(2024)Chen, Pan, Ye, Lu, and Xia]{chen2024each}
Ziyang Chen, Yongsheng Pan, Yiwen Ye, Mengkang Lu, and Yong Xia.
\newblock Each test image deserves a specific prompt: Continual test-time adaptation for 2d medical image segmentation.
\newblock In \emph{Proceedings of the IEEE/CVF conference on computer vision and pattern recognition}, pages 11184--11193, 2024.

\bibitem[Cioppa et~al.(2022)Cioppa, Giancola, Deliege, Kang, Zhou, Cheng, Ghanem, and Van~Droogenbroeck]{cioppa2022soccernet}
Anthony Cioppa, Silvio Giancola, Adrien Deliege, Le Kang, Xin Zhou, Zhiyu Cheng, Bernard Ghanem, and Marc Van~Droogenbroeck.
\newblock Soccernet-tracking: Multiple object tracking dataset and benchmark in soccer videos.
\newblock In \emph{2022 IEEE/CVF Conference on Computer Vision and Pattern Recognition Workshops (CVPRW)}. IEEE, 2022.

\bibitem[Cui et~al.(2023)Cui, Zeng, Zhao, Yang, Wu, and Wang]{cui2023sportsmot}
Yutao Cui, Chenkai Zeng, Xiaoyu Zhao, Yichun Yang, Gangshan Wu, and Limin Wang.
\newblock Sportsmot: A large multi-object tracking dataset in multiple sports scenes.
\newblock In \emph{Proceedings of the IEEE/CVF international conference on computer vision}, pages 9921--9931, 2023.

\bibitem[Dendorfer et~al.(2020)Dendorfer, Rezatofighi, Milan, Shi, Cremers, Reid, Roth, Schindler, and Leal-Taix{\'e}]{dendorfer2020mot20}
Patrick Dendorfer, Hamid Rezatofighi, Anton Milan, Javen Shi, Daniel Cremers, Ian Reid, Stefan Roth, Konrad Schindler, and Laura Leal-Taix{\'e}.
\newblock Mot20: A benchmark for multi object tracking in crowded scenes.
\newblock \emph{arXiv preprint arXiv:2003.09003}, 2020.

\bibitem[Gao and Wang(2023)]{gao2023memotr}
Ruopeng Gao and Limin Wang.
\newblock Memotr: Long-term memory-augmented transformer for multi-object tracking.
\newblock In \emph{Proceedings of the IEEE/CVF International Conference on Computer Vision}, pages 9901--9910, 2023.

\bibitem[Gao et~al.(2025)Gao, Qi, and Wang]{gao2025multiple}
Ruopeng Gao, Ji Qi, and Limin Wang.
\newblock Multiple object tracking as id prediction.
\newblock In \emph{Proceedings of the Computer Vision and Pattern Recognition Conference}, pages 27883--27893, 2025.

\bibitem[Guo et~al.(2026)Guo, Zhao, Wang, Hu, and Gao]{guo2026dual}
Wen Guo, Pengfei Zhao, Zongmeng Wang, Yufan Hu, and Junyu Gao.
\newblock Dual-level adaptation for multi-object tracking: Building test-time calibration from experience and intuition.
\newblock In \emph{Proceedings of the IEEE/CVF Conference on Computer Vision and Pattern Recognition}, pages 28190--28200, 2026.

\bibitem[Han et~al.(2025)Han, Yang, Wang, Li, Xu, Shou, and Zhang]{han2025dota}
Zongbo Han, Jialong Yang, Guangyu Wang, Junfan Li, Qianli Xu, Mike~Zheng Shou, and Changqing Zhang.
\newblock Dota: Distributional test-time adaptation of vision-language models.
\newblock \emph{Advances in Neural Information Processing Systems}, 2025.

\bibitem[Huang et~al.(2026)Huang, Pei, Wang, Liu, Li, and Wang]{huang2026motra}
Yuanzhou Huang, Songwei Pei, Shuhuai Wang, Bingfeng Liu, Qian Li, and Shangguang Wang.
\newblock Motra: Motion-aware target representation learning for end-to-end multi-object tracking.
\newblock In \emph{Proceedings of the Thirty-Fifth International Joint Conference on Artificial Intelligence, {IJCAI-26}}, 2026.

\bibitem[Kim et~al.(2025)Kim, Jeong, Kim, and Koh]{kim2025dual}
Hyunseop Kim, Juheon Jeong, Hanul Kim, and Yeong~Jun Koh.
\newblock Dual-path temporal decoder for end-to-end multi-object tracking.
\newblock \emph{Advances in Neural Information Processing Systems}, 2025.

\bibitem[Liu et~al.(2022)Liu, Li, Zhang, Yang, Qi, Su, Zhu, and Zhang]{liu2022dab}
Shilong Liu, Feng Li, Hao Zhang, Xiao Yang, Xianbiao Qi, Hang Su, Jun Zhu, and Lei Zhang.
\newblock Dab-detr: Dynamic anchor boxes are better queries for detr.
\newblock In \emph{International Conference on Learning Representations}, 2022.

\bibitem[Luiten et~al.(2021)Luiten, Osep, Dendorfer, Torr, Geiger, Leal-Taix{\'e}, and Leibe]{luiten2021hota}
Jonathon Luiten, Aljosa Osep, Patrick Dendorfer, Philip Torr, Andreas Geiger, Laura Leal-Taix{\'e}, and Bastian Leibe.
\newblock Hota: A higher order metric for evaluating multi-object tracking.
\newblock \emph{International journal of computer vision}, 129\penalty0 (2):\penalty0 548--578, 2021.

\bibitem[Meinhardt et~al.(2022)Meinhardt, Kirillov, Leal-Taixe, and Feichtenhofer]{meinhardt2022trackformer}
Tim Meinhardt, Alexander Kirillov, Laura Leal-Taixe, and Christoph Feichtenhofer.
\newblock Trackformer: Multi-object tracking with transformers.
\newblock In \emph{Proceedings of the IEEE/CVF conference on computer vision and pattern recognition}, pages 8844--8854, 2022.

\bibitem[Milan et~al.(2016)Milan, Leal-Taix{\'e}, Reid, Roth, and Schindler]{milan2016mot16}
Anton Milan, Laura Leal-Taix{\'e}, Ian Reid, Stefan Roth, and Konrad Schindler.
\newblock Mot16: A benchmark for multi-object tracking.
\newblock \emph{arXiv preprint arXiv:1603.00831}, 2016.

\bibitem[Pang et~al.(2021)Pang, Qiu, Li, Chen, Li, Darrell, and Yu]{pang2021quasi}
Jiangmiao Pang, Linlu Qiu, Xia Li, Haofeng Chen, Qi Li, Trevor Darrell, and Fisher Yu.
\newblock Quasi-dense similarity learning for multiple object tracking.
\newblock In \emph{Proceedings of the IEEE/CVF conference on computer vision and pattern recognition}, pages 164--173, 2021.

\bibitem[Ristani et~al.(2016)Ristani, Solera, Zou, Cucchiara, and Tomasi]{ristani2016performance}
Ergys Ristani, Francesco Solera, Roger Zou, Rita Cucchiara, and Carlo Tomasi.
\newblock Performance measures and a data set for multi-target, multi-camera tracking.
\newblock In \emph{European Conference on Computer Vision}, pages 17--35. Springer, 2016.

\bibitem[Ruan and Tang(2024)]{ruan2024fully}
Xiaoqian Ruan and Wei Tang.
\newblock Fully test-time adaptation for object detection.
\newblock In \emph{IEEE/CVF Conference on Computer Vision and Pattern Recognition Workshops (CVPRW)}, pages 1038--1047, 2024.

\bibitem[Scott et~al.(2024)Scott, Uchida, Ding, Umemoto, Bunker, Kobayashi, Koyama, Onishi, Kameda, and Fujii]{scott2024teamtrack}
Atom Scott, Ikuma Uchida, Ning Ding, Rikuhei Umemoto, Rory Bunker, Ren Kobayashi, Takeshi Koyama, Masaki Onishi, Yoshinari Kameda, and Keisuke Fujii.
\newblock Teamtrack: A dataset for multi-sport multi-object tracking in full-pitch videos.
\newblock In \emph{Proceedings-2024 IEEE/CVF Conference on Computer Vision and Pattern Recognition Workshops, CVPRW 2024}, pages 3357--3366. IEEE Computer Society, 2024.

\bibitem[Segu et~al.(2023)Segu, Schiele, and Yu]{segu2023darth}
Mattia Segu, Bernt Schiele, and Fisher Yu.
\newblock Darth: Holistic test-time adaptation for multiple object tracking.
\newblock In \emph{Proceedings of the IEEE/CVF International Conference on Computer Vision}, pages 9717--9727, 2023.

\bibitem[Segu et~al.(2025)Segu, Piccinelli, Li, Yang, Schiele, and Van~Gool]{segu2025samba}
Mattia Segu, Luigi Piccinelli, Siyuan Li, Yung-Hsu Yang, Bernt Schiele, and Luc Van~Gool.
\newblock Samba: Synchronized set-of-sequences modeling for multiple object tracking.
\newblock In \emph{The Thirteenth International Conference on Learning Representations}, pages 30057--30070. ICLR, 2025.

\bibitem[Shao et~al.(2026)Shao, Yang, Yu, Li, Guo, Yan, Wang, and Sun]{shao2026detection}
Yuqing Shao, Yuchen Yang, Rui Yu, Weilong Li, Xu Guo, Huaicheng Yan, Wei Wang, and Xiao Sun.
\newblock From detection to association: Learning discriminative object embeddings for multi-object tracking.
\newblock In \emph{Proceedings of the IEEE/CVF Conference on Computer Vision and Pattern Recognition}, pages 6878--6888, 2026.

\bibitem[Sun et~al.(2022)Sun, Cao, Jiang, Yuan, Bai, Kitani, and Luo]{sun2022dancetrack}
Peize Sun, Jinkun Cao, Yi Jiang, Zehuan Yuan, Song Bai, Kris Kitani, and Ping Luo.
\newblock Dancetrack: Multi-object tracking in uniform appearance and diverse motion.
\newblock In \emph{Proceedings of the IEEE/CVF conference on computer vision and pattern recognition}, pages 20993--21002, 2022.

\bibitem[Tang et~al.(2019)Tang, Naphade, Liu, Yang, Birchfield, Wang, Kumar, Anastasiu, and Hwang]{tang2019cityflow}
Zheng Tang, Milind Naphade, Ming-Yu Liu, Xiaodong Yang, Stan Birchfield, Shuo Wang, Ratnesh Kumar, David Anastasiu, and Jenq-Neng Hwang.
\newblock Cityflow: A city-scale benchmark for multi-target multi-camera vehicle tracking and re-identification.
\newblock In \emph{Proceedings of the IEEE/CVF conference on computer vision and pattern recognition}, pages 8797--8806, 2019.

\bibitem[Wang et~al.(2021)Wang, Shelhamer, Liu, Olshausen, and Darrell]{wang2021tent}
Dequan Wang, Evan Shelhamer, Shaoteng Liu, Bruno Olshausen, and Trevor Darrell.
\newblock Tent: Fully test-time adaptation by entropy minimization.
\newblock In \emph{International Conference on Learning Representations}, 2021.

\bibitem[Wang et~al.(2025)Wang, Wang, Cao, Chen, and Li]{wang2025motr}
Peng Wang, Yongcai Wang, Hualong Cao, Wang Chen, and Deying Li.
\newblock La-motr: End-to-end multi-object tracking by learnable association.
\newblock In \emph{Proceedings of the IEEE/CVF International Conference on Computer Vision}, pages 12438--12448, 2025.

\bibitem[Wang et~al.(2022)Wang, Fink, Van~Gool, and Dai]{wang2022continual}
Qin Wang, Olga Fink, Luc Van~Gool, and Dengxin Dai.
\newblock Continual test-time domain adaptation.
\newblock In \emph{Proceedings of the IEEE/CVF conference on computer vision and pattern recognition}, pages 7201--7211, 2022.

\bibitem[Wojke et~al.(2017)Wojke, Bewley, and Paulus]{wojke2017simple}
Nicolai Wojke, Alex Bewley, and Dietrich Paulus.
\newblock Simple online and realtime tracking with a deep association metric.
\newblock In \emph{2017 IEEE international conference on image processing (ICIP)}, pages 3645--3649. IEEE, 2017.

\bibitem[Yan et~al.(2025)Yan, Luo, Zhong, Gan, and Ma]{luo2025co}
Feng Yan, Weixin Luo, Yujie Zhong, Yiyang Gan, and Lin Ma.
\newblock Co-mot: Boosting end-to-end transformer-based multi-object tracking via coopetition label assignment and shadow sets.
\newblock In \emph{The Thirteenth International Conference on Learning Representations}, 2025.

\bibitem[Yang et~al.(2026)Yang, Rao, Wu, and Xie]{yang2026soccermaster}
Haolin Yang, Jiayuan Rao, Haoning Wu, and Weidi Xie.
\newblock Soccermaster: A vision foundation model for soccer understanding.
\newblock In \emph{Proceedings of the IEEE/CVF Conference on Computer Vision and Pattern Recognition}, 2026.

\bibitem[Yang and Agam(2025)]{yang2025motion}
Xu Yang and Gady Agam.
\newblock Motion-aware transformer for multi-object tracking.
\newblock \emph{arXiv preprint arXiv:2509.21715}, 2025.

\bibitem[Yu et~al.(2020)Yu, Chen, Wang, Xian, Chen, Liu, Madhavan, and Darrell]{yu2020bdd100k}
Fisher Yu, Haofeng Chen, Xin Wang, Wenqi Xian, Yingying Chen, Fangchen Liu, Vashisht Madhavan, and Trevor Darrell.
\newblock Bdd100k: A diverse driving dataset for heterogeneous multitask learning.
\newblock In \emph{Proceedings of the IEEE/CVF conference on computer vision and pattern recognition}, pages 2636--2645, 2020.

\bibitem[Zeng et~al.(2022)Zeng, Dong, Zhang, Wang, Zhang, and Wei]{zeng2022motr}
Fangao Zeng, Bin Dong, Yuang Zhang, Tiancai Wang, Xiangyu Zhang, and Yichen Wei.
\newblock Motr: End-to-end multiple-object tracking with transformer.
\newblock In \emph{European conference on computer vision}, pages 659--675. Springer, 2022.

\bibitem[Zeng et~al.(2025)Zeng, Huang, and Pei]{zeng2025tgformer}
Rui Zeng, Yuanzhou Huang, and Songwei Pei.
\newblock Tgformer: Transformer with track query group for multi-object tracking.
\newblock In \emph{Proceedings of the AAAI Conference on Artificial Intelligence}, pages 9824--9832, 2025.

\bibitem[Zhang et~al.(2021)Zhang, Wang, Wang, Zeng, and Liu]{zhang2021fairmot}
Yifu Zhang, Chunyu Wang, Xinggang Wang, Wenjun Zeng, and Wenyu Liu.
\newblock Fairmot: On the fairness of detection and re-identification in multiple object tracking.
\newblock \emph{International journal of computer vision}, 129\penalty0 (11):\penalty0 3069--3087, 2021.

\bibitem[Zhang et~al.(2022)Zhang, Sun, Jiang, Yu, Weng, Yuan, Luo, Liu, and Wang]{zhang2022bytetrack}
Yifu Zhang, Peize Sun, Yi Jiang, Dongdong Yu, Fucheng Weng, Zehuan Yuan, Ping Luo, Wenyu Liu, and Xinggang Wang.
\newblock Bytetrack: Multi-object tracking by associating every detection box.
\newblock In \emph{European conference on computer vision}, pages 1--21. Springer, 2022.

\bibitem[Zhang et~al.(2023)Zhang, Wang, and Zhang]{zhang2023motrv2}
Yuang Zhang, Tiancai Wang, and Xiangyu Zhang.
\newblock Motrv2: Bootstrapping end-to-end multi-object tracking by pretrained object detectors.
\newblock In \emph{Proceedings of the IEEE/CVF conference on computer vision and pattern recognition}, pages 22056--22065, 2023.

\bibitem[Zhu et~al.(2018)Zhu, Wen, Bian, Ling, and Hu]{zhu2018vision}
Pengfei Zhu, Longyin Wen, Xiao Bian, Haibin Ling, and Qinghua Hu.
\newblock Vision meets drones: A challenge.
\newblock \emph{arXiv preprint arXiv:1804.07437}, 2018.

\bibitem[Zhu et~al.(2021)Zhu, Su, Lu, Li, Wang, and Dai]{zhu2021deformable}
Xizhou Zhu, Weijie Su, Lewei Lu, Bin Li, Xiaogang Wang, and Jifeng Dai.
\newblock Deformable detr: Deformable transformers for end-to-end object detection.
\newblock In \emph{International Conference on Learning Representations}, 2021.

\end{thebibliography}
}

\clearpage
\appendix

\twocolumn[
{
    \centering
    {\LARGE\bfseries Supplementary Material\par}
    \vspace{0.2in}
    {\large\bfseries
    Calibrating Track-Query Updates at Test Time for Multi-Object Tracking\par}
    \vspace{0.5in}
}
]

\section{Introduction}
\label{sec:supp_introduction}

This supplementary material provides further implementation details, analyses, and experimental results supporting the main paper.
We first report the hyperparameters used for test-set evaluation and extend the query-update analysis to SambaMOTR.
We then examine the robustness of the query-space analysis, controlled variants, the effects of TQC on track-query updates, identity-switch counts, and performance on association-hard sequences.
Finally, we report hyperparameter sensitivity, cross-dataset hyperparameter transfer, additional dataset evaluations, and computational complexity.

\section{Further Implementation Details}
\label{sec:supp_implementation_details}

TQC selects up to $K=5$ candidate tracks from other active tracks with positive bounding-box IoU.
TQC is applied to continuing tracks whose queries are updated by the tracker at the current frame, while newly initialized tracks can serve as candidates but are not themselves calibrated.
For the nine tracker and dataset settings in the main paper, we select $(\beta,\rho)$ using HOTA~\cite{luiten2021hota} on the corresponding validation set and fix them for test-set evaluation.
For the correction-direction controls, we use the parameters selected for full TQC on the same validation sets.
These controls preserve the resulting correction magnitudes and are not tuned separately.
Each reported tracking result is obtained using a fixed pretrained checkpoint.
TQC introduces no stochastic operations during inference.
The selected values are summarized in Table~\ref{tab:supp_selected_hyperparameters}.
The sensitivity of TQC to these parameters is examined later in the supplementary material.

\paragraph{CO-MOT Implementation.}
For CO-MOT~\cite{luo2025co}, TQC operates on individual shadow queries rather than tracked identities.
Candidates are selected using IoU between their predicted boxes, with the top-$K$ counted over shadow queries and those belonging to the same identity excluded.
Calibration is applied only to continuing shadow queries selected for updating by the tracker, based on their individual scores (threshold $0.5$).

\paragraph{DualMOT Evaluation.}
We evaluate DualMOT~\cite{kim2025dual} using its publicly released tracking checkpoint trained on CrowdHuman and DanceTrack train.
Without confidence-guided update suppression, we reproduce the validation result reported for this configuration ($67.85$ vs.\ $67.9$ HOTA) and obtain $72.88$ HOTA on the test set.
With suppression enabled (threshold $0.4$), we obtain $67.42$ HOTA on validation and $72.60$ on test, below the reported $69.1$ and $74.1$, respectively.
We therefore use the reproduced baseline without suppression.

\begin{table}[t]
\centering
\caption{
\textbf{Hyperparameters used for test-set evaluation.}
The parameters are selected using HOTA on the corresponding validation set and fixed for test-set evaluation.
$^\dagger$ DualMOT uses the publicly released checkpoint without confidence-guided update suppression.
}
\small
\setlength{\tabcolsep}{4.0pt}
\begin{tabular}{@{}llcc@{}}
\toprule
Dataset & Tracker & $\beta$ & $\rho$ \\
\midrule
DanceTrack
& MOTR~\cite{zeng2022motr} & 1.50 & 0.15 \\
& MeMOTR-Deformable~\cite{gao2023memotr} & 1.25 & 0.05 \\
& CO-MOT~\cite{luo2025co} & 1.00 & 0.10 \\
& MeMOTR-DAB~\cite{gao2023memotr} & 1.75 & 0.15 \\
& SambaMOTR-DAB~\cite{segu2025samba} & 0.75 & 0.50 \\
& DualMOT$^\dagger$~\cite{kim2025dual} & 2.00 & 0.25 \\
\midrule
SportsMOT
& MeMOTR-Deformable & 1.50 & 0.15 \\
& MeMOTR-DAB & 0.75 & 0.10 \\
& SambaMOTR-DAB & 0.50 & 0.75 \\
\bottomrule
\end{tabular}
\label{tab:supp_selected_hyperparameters}
\end{table}

\section{Additional Query-Update Analysis with SambaMOTR}
\label{sec:supp_samba_analysis}

The main paper analyzes track-query updates using MeMOTR-DAB~\cite{gao2023memotr,liu2022dab}.
Here, we repeat the same analysis using SambaMOTR-DAB~\cite{segu2025samba} on the DanceTrack~\cite{sun2022dancetrack} and SportsMOT~\cite{cui2023sportsmot} validation sets.
For analysis, we consider true-positive track states and exclude states for which an identity switch (IDSW) occurs at frame $t$.
Among the remaining states, we label a state at frame $t$ as pre-IDSW if an IDSW occurs for the same ground-truth trajectory within the next $H$ frames, and as stable otherwise.
We use the same future window $H=3$ as in the main analysis.
These labels are used only for analysis and are not used by TQC.

\begin{figure*}[t]
\centering
\setlength{\tabcolsep}{2pt}
\begin{tabular}{ccc}
\includegraphics[width=0.325\textwidth]{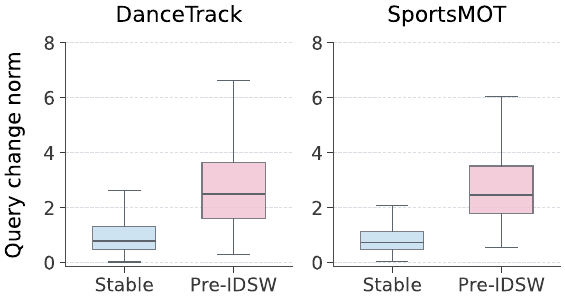} &
\includegraphics[width=0.325\textwidth]{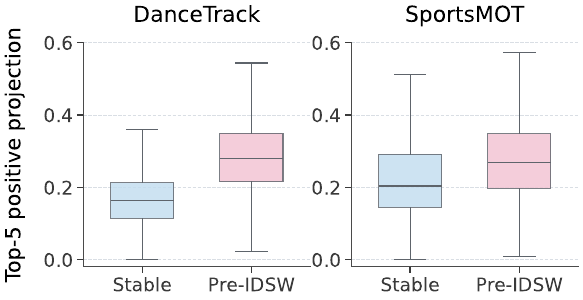} &
\includegraphics[width=0.325\textwidth]{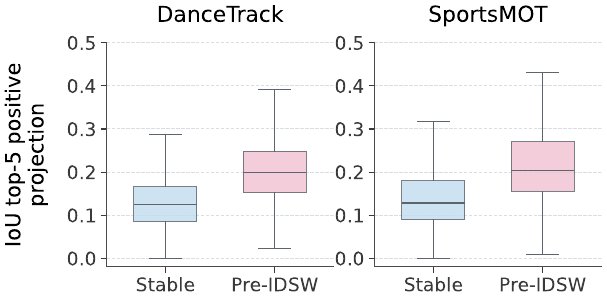} \\
\small (a) & \small (b) & \small (c)
\end{tabular}
\caption{
\textbf{Distributions of query-update statistics before IDSWs with SambaMOTR-DAB.}
(a) Norm of the frame-to-frame change in propagated track-query representations.
(b) Average of the five largest normalized positive projections over all other active tracks.
(c) Average normalized positive projection over IoU top-5 nearby active tracks.
}
\label{fig:supp_samba_query_update_distributions}
\end{figure*}

\subsection{Propagated Track Queries Change More Abruptly Before IDSWs}

For track $i$, let $q_i^t$ denote the propagated track-query representation at frame $t$, before incorporating current-frame features.
We define its frame-to-frame change as:
\begin{equation}
\Delta q_i^t = q_i^{t+1} - q_i^t .
\label{eq:supp_samba_delta_q}
\end{equation}

Fig.~\ref{fig:supp_samba_query_update_distributions}(a) compares the distributions of $\lVert \Delta q_i^t \rVert_2$ for stable and pre-IDSW states.
On DanceTrack, the mean norm increases from $1.046$ for stable states to $2.883$ for pre-IDSW states, with an AUROC of $0.877$.
On SportsMOT, the corresponding mean norms are $0.945$ and $3.055$, with an AUROC of $0.932$.
These results show that propagated track-query representations also change more abruptly before IDSWs in SambaMOTR-DAB.

\subsection{Query Updates Show Stronger Positive Projections Toward Other Tracks Before IDSWs}

We next examine whether the current-frame query update contains components directed toward other active tracks.
Let $o_i^t$ denote the tracker output representation after incorporating current-frame image features.
We define the current-frame query update as:
\begin{equation}
d_i^t = o_i^t - q_i^t .
\label{eq:supp_samba_update_direction}
\end{equation}

For another active track $j$, we define the direction from track $i$ to track $j$ in the query representation space as:
\begin{equation}
u_{ij}^t =
\frac{q_j^t - q_i^t}
{\lVert q_j^t - q_i^t\rVert_2} .
\label{eq:supp_samba_candidate_direction}
\end{equation}

We measure the alignment between the query update and this direction using the normalized positive projection:
\begin{equation}
\bar{p}_{ij}^t =
\frac{
\left[(d_i^t)^\top u_{ij}^t\right]_+
}{
\lVert d_i^t\rVert_2
},
\label{eq:supp_samba_normalized_positive_projection}
\end{equation}
where $[\cdot]_+ = \max(\cdot,0)$.
For each track $i$, we compute $\bar{p}_{ij}^t$ over all other active tracks and average the five largest values.

Fig.~\ref{fig:supp_samba_query_update_distributions}(b) shows that this statistic is larger for pre-IDSW states than for stable states on both datasets.
On DanceTrack, the mean score increases from $0.166$ to $0.284$, with an AUROC of $0.832$.
On SportsMOT, the corresponding means are $0.226$ and $0.276$, with an AUROC of $0.640$.
The separation is weaker on SportsMOT, although the mean remains higher for pre-IDSW states.
Thus, the tendency toward stronger alignment with directions toward other active tracks also appears in SambaMOTR-DAB, while its strength varies across datasets.

\begin{table*}[t]
\centering
\caption{
\textbf{AUROC of normalized positive projections after covariance whitening.}
Results are reported on six validation settings.
Brackets indicate $95\%$ confidence intervals obtained from $1{,}000$ sequence-level bootstrap resamples.
}
\small
\setlength{\tabcolsep}{7.0pt}
\begin{tabular}{@{}lcccc@{}}
\toprule
& \multicolumn{2}{c}{All other active tracks}
& \multicolumn{2}{c}{IoU top-5 candidates} \\
\cmidrule(lr){2-3}
\cmidrule(l){4-5}
Tracker
& AUROC
& $95\%$ CI
& AUROC
& $95\%$ CI \\
\midrule
\multicolumn{5}{@{}l}{\textit{DanceTrack}} \\
MeMOTR-Deformable
& 0.680 & [0.625, 0.735]
& 0.657 & [0.615, 0.705] \\
MeMOTR-DAB
& 0.699 & [0.661, 0.736]
& 0.682 & [0.647, 0.717] \\
SambaMOTR-DAB
& 0.691 & [0.649, 0.734]
& 0.666 & [0.635, 0.697] \\
\midrule
\multicolumn{5}{@{}l}{\textit{SportsMOT}} \\
MeMOTR-Deformable
& 0.686 & [0.635, 0.732]
& 0.698 & [0.651, 0.739] \\
MeMOTR-DAB
& 0.676 & [0.611, 0.736]
& 0.689 & [0.630, 0.743] \\
SambaMOTR-DAB
& 0.720 & [0.676, 0.765]
& 0.741 & [0.701, 0.783] \\
\bottomrule
\end{tabular}
\label{tab:supp_geometry_robustness}
\end{table*}

\begin{table}[t]
\centering
\caption{
\textbf{Coverage of the future IDSW counterpart by the candidate set used by TQC.}
}
\small
\setlength{\tabcolsep}{4.0pt}
\begin{tabular}{@{}lcc@{}}
\toprule
Candidate set
& Track states
& IDSW events \\
\midrule
IoU top-5
& $865/873$ ($99.08\%$)
& $405/408$ ($99.26\%$) \\
\bottomrule
\end{tabular}
\label{tab:supp_tqc_candidate_coverage}
\end{table}

\begin{table}[t]
\centering
\caption{
\textbf{Effect of TQC on the update component directed toward the IDSW counterpart.}}
\small
\setlength{\tabcolsep}{3.5pt}
\begin{tabular}{@{}lr@{}}
\toprule
Measure & Result \\
\midrule
Nonzero counterpart correction term
& $767/865$ ($88.67\%$) \\
Counterpart projection decreased
& $752/767$ ($98.04\%$) \\
Median update-norm retention
& $0.971$ \\
Median counterpart-projection retention
& $0.378$ \\
\bottomrule
\end{tabular}
\label{tab:supp_tqc_fixed_state_effect}
\end{table}

\begin{table*}[t]
\centering
\caption{
\textbf{Effect of TQC on candidate-directed projections immediately after TQC and after subsequent query processing.}
Each percentage is computed over states for which the corresponding paired comparison is defined.
Retention values are state-wise ratios reported as medians and interquartile ranges.
}
\small
\setlength{\tabcolsep}{5.0pt}
\begin{tabular}{@{}lccc@{}}
\toprule
Representation
& Projection decreased
& \shortstack{Median projection\\retention [IQR]}
& \shortstack{Median update-norm\\retention [IQR]} \\
\midrule
Tracker output after TQC
& $98.38\%$
& $0.317$ $[0.109,0.467]$
& $0.980$ $[0.970,0.990]$ \\
Track query used for propagation
& $97.21\%$
& $0.235$ $[0.002,0.503]$
& $0.792$ $[0.623,0.920]$ \\
\bottomrule
\end{tabular}
\label{tab:supp_tqc_propagated_effect}
\end{table*}


\begin{figure}[t]
\centering
\includegraphics[width=\linewidth]{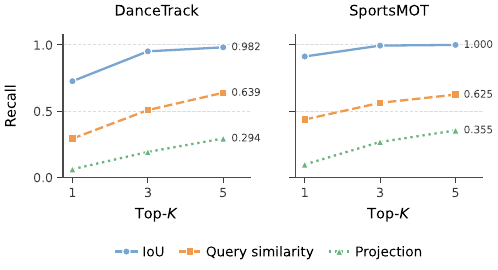}
\caption{
\textbf{Candidate retrieval recall for future IDSW counterparts with SambaMOTR-DAB.}
For each pre-IDSW state in which the counterpart track is already active, IoU-based retrieval considers only tracks with positive IoU, as in TQC.
Query cosine similarity and the normalized positive projection rank all other active tracks.
}
\label{fig:supp_samba_candidate_retrieval_recall}
\end{figure}

\subsection{IoU Retrieves IDSW Counterpart Tracks}

Finally, we compare bounding-box IoU, query cosine similarity, and the normalized positive projection as criteria for retrieving the counterpart track involved in a subsequent IDSW.
We evaluate retrieval only for pre-IDSW states in which the counterpart track is already active.
For IoU-based retrieval, we consider only tracks with positive IoU, as in TQC, while the other two criteria rank all other active tracks.

Fig.~\ref{fig:supp_samba_candidate_retrieval_recall} reports top-$K$ recall for retrieving the future IDSW counterpart.
IoU achieves the highest top-$K$ recall among the three criteria.
In particular, IoU top-5 candidates include the counterpart track in $98.2\%$ of evaluated states on DanceTrack and $100.0\%$ on SportsMOT.
These results are conditional on the counterpart being active and do not represent coverage of all IDSW events.

We then compute the average normalized positive projection over the IoU top-5 candidates.
As shown in Fig.~\ref{fig:supp_samba_query_update_distributions}(c), this score is larger for pre-IDSW states than for stable states on both datasets.
On DanceTrack, the mean score increases from $0.128$ to $0.205$, with an AUROC of $0.791$.
On SportsMOT, the corresponding means are $0.142$ and $0.213$, with an AUROC of $0.747$.
These results show that query updates are more strongly aligned with directions toward spatially nearby active tracks before IDSWs.

Overall, the SambaMOTR-DAB results support the three findings observed with MeMOTR-DAB.
Pre-IDSW states exhibit larger propagated-query changes, stronger update alignment with directions toward other active tracks, and high recall of future IDSW counterparts using IoU.
Although the separation over all active tracks is weaker on SportsMOT, restricting the analysis to IoU-based nearby candidates yields clearer separation and supports the use of IoU for candidate retrieval in TQC.


\section{Robustness of the Query-Space Analysis}
\label{sec:supp_geometry_robustness}

The main paper shows that current-frame query updates have larger positive projections toward other active tracks before IDSWs.
However, Euclidean projections depend on the coordinate system of the learned query representations.
We therefore examine whether this separation persists after covariance whitening.

We analyze baseline trajectories of MeMOTR-Deformable, MeMOTR-DAB, and SambaMOTR-DAB on the DanceTrack and SportsMOT validation sets.
We use the same definitions of pre-IDSW and stable states as in the main paper, with $H=3$.
We estimate the covariance matrix $\Sigma$ separately for each setting from all baseline validation track-query representations without using ground-truth labels.
We apply covariance whitening to both the propagated track queries and tracker outputs:
\begin{equation}
T=\Sigma^{-1/2}, \qquad
q_i^{\prime t}=Tq_i^t, \qquad
o_i^{\prime t}=To_i^t.
\end{equation}

We recompute the normalized positive projections in the transformed space and evaluate their ability to distinguish pre-IDSW from stable states using AUROC.
As in the main paper, we consider the five largest projections over all other active tracks and the average projection over IoU top-5 candidates.

As shown in Table~\ref{tab:supp_geometry_robustness}, the separation between pre-IDSW and stable states persists after covariance whitening across all six settings.
The AUROC ranges from $0.676$ to $0.720$ over all other active tracks and from $0.657$ to $0.741$ over IoU top-5 candidates.
The lower bounds of the $95\%$ confidence intervals exceed $0.5$ in every setting.

We further apply a random invertible linear transformation (condition number $96.2$) to the query representations and tracker outputs and recompute the covariance matrix after the transformation.
Averaged over the six settings, the AUROC over IoU top-5 candidates is $0.689$ before and $0.690$ after the transformation, with a maximum per-setting difference of $0.009$.
These results are consistent with the invariance of the covariance-whitened metric under invertible linear transformations.

Finally, we compare pre-IDSW and stable states within the same frame to reduce the influence of differences in frame-level conditions.
Under covariance whitening, the AUROC ranges from $0.600$ to $0.716$ over all other active tracks and from $0.590$ to $0.730$ over IoU top-5 candidates.
The separation therefore remains observable even when comparisons are restricted to states within the same frame.

\begin{table*}[t]
\centering
\caption{
\textbf{Controlled variants on six validation settings.}
Values are changes from the corresponding baseline.
Direction variants preserve the TQC correction norm, and random directions are averaged over eight seeds.
Norm-preserving correction restores the original update norm, update damping reduces the magnitude of the current-frame query update by the magnitude of the TQC correction, and uniform damping scales continuing-track updates by $\gamma$.
}
\small
\renewcommand{\arraystretch}{0.90}
\setlength{\tabcolsep}{2.5pt}
\begin{tabular}{@{}lrrrrrrccc@{}}
\toprule
& \multicolumn{6}{c}{\(\Delta\)HOTA}
& \multicolumn{3}{c}{Average} \\
\cmidrule(lr){2-7}
\cmidrule(l){8-10}
Variant
& \shortstack{MeMOTR\\DAB DT}
& \shortstack{MeMOTR\\Def DT}
& \shortstack{SambaMOTR\\DAB DT}
& \shortstack{MeMOTR\\DAB SP}
& \shortstack{MeMOTR\\Def SP}
& \shortstack{SambaMOTR\\DAB SP}
& \(\Delta\)HOTA
& \(\Delta\)AssA
& \(\Delta\)IDF1 \\
\midrule
\textbf{Full TQC}
& +0.85 & +0.96 & +0.62 & +0.50 & +1.22 & +0.09
& \textbf{+0.71} & \textbf{+1.06} & \textbf{+0.88} \\
\midrule

\multicolumn{10}{@{}l}{\textit{Correction direction}} \\
Reversed correction
& -2.88 & -0.64 & -0.52 & -0.41 & -0.81 & -0.20
& -0.91 & -1.28 & -1.36 \\
Random directions
& -0.48 & +0.27 & -0.10 & +0.13 & +0.11 & -0.15
& -0.04 & -0.05 & -0.07 \\
Orthogonal directions
& -0.35 & -0.31 & +0.77 & +0.06 & -0.22 & -0.29
& -0.06 & -0.09 & +0.02 \\
Non-overlapping-track directions
& +0.50 & +0.39 & +0.08 & -0.35 & +0.44 & -0.44
& +0.10 & +0.15 & +0.01 \\
\midrule

\multicolumn{10}{@{}l}{\textit{Calibration strategy}} \\
Norm-preserving correction
& +0.44 & +0.82 & +0.14 & -0.36 & +0.77 & -0.02
& +0.30 & +0.35 & +0.33 \\
Update damping
& +0.99 & -0.20 & +0.19 & +0.38 & +0.26 & +0.11
& +0.29 & +0.48 & +0.39 \\
Uniform damping (\(\gamma=0.95\))
& +0.08 & +0.98 & +0.46 & +0.40 & +0.02 & -0.12
& +0.30 & +0.45 & +0.42 \\
Uniform damping (\(\gamma=0.75\))
& +0.39 & -0.72 & +0.33 & +0.43 & +0.19 & +0.04
& +0.11 & +0.10 & +0.19 \\
Uniform damping (\(\gamma=0.50\))
& -0.65 & -6.18 & -0.22 & -0.01 & -0.38 & -0.55
& -1.33 & -1.90 & -1.70 \\
\midrule

\multicolumn{10}{@{}l}{\textit{Candidate weighting}} \\
w/o IoU weights
& -0.51 & +0.31 & +0.32 & +0.24 & +0.76 & -0.22
& +0.15 & +0.25 & +0.21 \\
Projection weights
& -0.48 & +0.21 & +0.37 & +0.29 & +0.59 & -0.17
& +0.13 & +0.24 & +0.04 \\
\midrule

\multicolumn{10}{@{}l}{\textit{Correction bound}} \\
w/o bound
& -4.81 & -5.20 & +0.43 & -0.16 & -7.78 & +0.09
& -2.91 & -3.70 & -3.33 \\
\midrule

\multicolumn{10}{@{}l}{\textit{Number of candidates}} \\
\(K=3\)
& +0.42 & +1.09 & +0.48 & +0.41 & +1.20 & -0.01
& +0.60 & +0.89 & +0.70 \\
\(K=10\)
& +0.56 & +0.97 & +0.91 & +0.46 & +1.14 & +0.09
& +0.69 & +1.03 & +0.77 \\
\bottomrule
\end{tabular}
\label{tab:supp_controls_six_settings}
\end{table*}

\section{Per-Setting Results for Controlled Variants}
\label{sec:supp_controls}

The main paper summarizes the average effects of controlled variants across six validation settings.
Here, we report the corresponding HOTA changes for each tracker and dataset setting in Table~\ref{tab:supp_controls_six_settings}.


\section{Effects of TQC on Track-Query Updates}
\label{sec:supp_tqc_update_effects}

The main paper summarizes how TQC changes candidate-directed update components.
Here, we provide complementary analyses of the specific counterpart involved in a subsequent IDSW and more detailed statistics for states encountered during TQC inference.
We evaluate counterpart coverage, the immediate effect of the bounded correction on projections toward the counterpart, and whether the reduction persists in the track-query representation used for subsequent propagation.

Unless otherwise stated, we use MeMOTR-DAB on the DanceTrack validation set and the same future window $H=3$ as in the main analysis.
We exclude states at which an IDSW occurs in the current frame.

\subsection{Coverage of Future IDSW Counterparts by TQC Candidates}
\label{sec:supp_tqc_candidate_coverage}

The main paper reports counterpart coverage under the positive-IoU candidate-selection rule used by TQC.
Here, we provide the corresponding counts and additionally report coverage over IDSW events.
An IDSW event is covered if its counterpart is selected within the preceding $H=3$ frames.

Table~\ref{tab:supp_tqc_candidate_coverage} shows that the future IDSW counterpart is included in the TQC candidate set for approximately $99\%$ of both evaluated track states and IDSW events.

\subsection{Reduction of Counterpart-Directed Update Components}
\label{sec:supp_tqc_counterpart_reduction}

Including the IDSW counterpart in the candidate set does not establish that the final bounded correction reduces the update component directed toward that track.
We therefore hold each tracker state fixed and compare the current-frame query update before and after applying TQC.
This comparison includes correction terms from all selected candidates and the norm bound defined in the main paper.

For track $i$, the current-frame query update is:
\begin{equation}
d_i^t = o_i^t - q_i^t ,
\label{eq:supp_tqc_update}
\end{equation}
where $q_i^t$ is the propagated track-query representation and $o_i^t$ is the tracker output representation.
For another active track $j$, the direction from track $i$ to track $j$ in the query representation space is:
\begin{equation}
u_{ij}^t =
\frac{q_j^t-q_i^t}
{\lVert q_j^t-q_i^t\rVert_2}.
\label{eq:supp_tqc_candidate_direction}
\end{equation}

We denote the calibrated tracker output by $\tilde{o}_i^t$ and define the corresponding query update as:
\begin{equation}
\tilde{d}_i^t = \tilde{o}_i^t-q_i^t .
\label{eq:supp_tqc_calibrated_update}
\end{equation}
For the IDSW counterpart $j^\star$, we compute the retained update norm and retained positive projection as:
\begin{equation}
R_{\mathrm{upd},i}^t =
\frac{\lVert \tilde{d}_i^t\rVert_2}
{\lVert d_i^t\rVert_2},
\qquad
R_{\mathrm{proj},ij^\star}^t =
\frac{
\left[(\tilde{d}_i^t)^\top u_{ij^\star}^t\right]_+
}{
\left[(d_i^t)^\top u_{ij^\star}^t\right]_+
}.
\label{eq:supp_tqc_retention}
\end{equation}

Unlike the normalized projection used to compare stable and pre-IDSW states in the main analysis, $R_{\mathrm{proj},ij^\star}^t$ measures the retention of the absolute positive projection after calibration.

Table~\ref{tab:supp_tqc_fixed_state_effect} summarizes the fixed-state comparison.
Among the states in which the counterpart is selected and the source track is eligible for TQC, the counterpart contributes a nonzero individual correction term in most cases.
For these states, the final correction decreases the positive projection toward the counterpart in nearly all cases.
The median retention values show that TQC preserves most of the update magnitude while substantially reducing the counterpart-directed projection.

The counterpart-directed projection is completely eliminated in $174$ states ($22.69\%$).
Because candidate directions need not be orthogonal, correction terms for different candidates may interact in the query representation space.

\subsection{Persistence of Projection Reduction After Subsequent Query Processing}
\label{sec:supp_tqc_propagated_effect}

The preceding fixed-state analysis measures the immediate effect of TQC on pre-IDSW states recorded from the baseline trajectory.
We next examine whether the same reduction is observed in states encountered during TQC inference and remains in the track-query representation used for subsequent propagation.

For each state encountered during TQC inference, we compare two branches initialized from the same tracker state.
The first applies TQC, whereas the second disables only the current-frame correction.
This paired comparison isolates the local effect of the current correction and avoids comparing independently rolled-out baseline and TQC trajectories.

Unlike the preceding analysis, which focuses on the specific counterpart involved in a subsequent IDSW, this analysis considers the candidate directions selected by TQC, regardless of whether an IDSW subsequently occurs.
We evaluate all $25$ DanceTrack validation sequences.

At each representation stage, we compute the average positive projection over the candidate directions selected by TQC.
Projection retention is the ratio of this value with TQC to that obtained from the paired branch with the current-frame correction disabled.
Update-norm retention is defined analogously using the norm of the corresponding representation change.
We report the median and interquartile range of these state-wise retention ratios over states for which the corresponding comparison is defined.

As shown in Table~\ref{tab:supp_tqc_propagated_effect}, the candidate-directed projection decreases in $98.38\%$ of states immediately after TQC.
The median projection retention is $0.317$, whereas the median update-norm retention is $0.980$, showing that TQC substantially reduces the selected directional component while retaining most of the immediate update magnitude.

In the track-query representation used for subsequent propagation, the candidate-directed projection remains lower in $97.21\%$ of states.
The decrease is observed in all $25$ validation sequences. 
Moreover, among the states for which both comparisons are defined, $95.65\%$ show a projection decrease at both stages.
Thus, the reduction in candidate-directed projection is largely preserved after subsequent query processing.


\begin{table}[t]
\centering
\caption{
\textbf{Identity-switch counts on the DanceTrack test set.}
Lower values are better.
$\Delta$ denotes TQC $-$ Base.
}
\small
\setlength{\tabcolsep}{3.5pt}
\begin{tabular}{@{}lrrr@{}}
\toprule
Tracker
& Base
& TQC
& $\Delta$ \\
\midrule
\multicolumn{4}{@{}l}{\textit{Deformable:}} \\
MOTR
& 2{,}383 & \textbf{2{,}330} & -53 \\
MeMOTR
& 1{,}477 & \textbf{1{,}378} & -99 \\
CO-MOT
& 1{,}254 & \textbf{1{,}184} & -70 \\
\midrule
\multicolumn{4}{@{}l}{\textit{DAB:}} \\
MeMOTR
& 1{,}115 & \textbf{1{,}106} & -9 \\
SambaMOTR
& \textbf{1{,}936} & 1{,}995 & +59 \\
\midrule
DualMOT$^\dagger$
& 770 & \textbf{767} & -3 \\
\bottomrule
\end{tabular}
\label{tab:test_error_counts_dancetrack}
\end{table}

\begin{table}[t]
\centering
\caption{
\textbf{Identity-switch counts on the SportsMOT test set.}
Lower values are better.
$\Delta$ denotes TQC $-$ Base.
}
\small
\setlength{\tabcolsep}{3.5pt}
\begin{tabular}{@{}lrrr@{}}
\toprule
Tracker
& Base
& TQC
& $\Delta$ \\
\midrule
\multicolumn{4}{@{}l}{\textit{Deformable:}} \\
MeMOTR
& 3{,}359 & \textbf{3{,}238} & -121 \\
\midrule
\multicolumn{4}{@{}l}{\textit{DAB:}} \\
MeMOTR
& 3{,}393 & \textbf{3{,}292} & -101 \\
SambaMOTR
& 3{,}129 & \textbf{2{,}835} & -294 \\
\bottomrule
\end{tabular}
\label{tab:test_error_counts_sportsmot}
\end{table}

\section{Identity Switch Counts}
\label{sec:supp_additional_identity_association}

The preceding sections examine how TQC changes track-query updates at individual tracker states.
The main paper summarizes the overall trend in IDSW counts.
Here, we report the complete IDSW counts for all evaluated tracker and dataset settings.

Tables~\ref{tab:test_error_counts_dancetrack} and~\ref{tab:test_error_counts_sportsmot} report the results.
TQC reduces IDSWs in eight of the nine evaluated settings.

IDSW counts record individual identity switches, whereas AssA and IDF1 measure association quality over the resulting trajectories.
These quantities therefore need not vary monotonically because event counts do not capture the duration of correct or incorrect identity associations.
IDSW increases only for SambaMOTR-DAB on DanceTrack, where HOTA, AssA, and IDF1 nevertheless improve.

\section{Additional Results on Association-Hard Sequences}
\label{sec:supp_association_hard}

We further evaluate TQC on validation sequences with challenging identity association.
For each tracker, we select the ten sequences with the largest values of baseline $\mathrm{DetA}-\mathrm{AssA}$, indicating a large gap between detection and association quality.
We also examine whether the results are preserved under an alternative selection criterion based on baseline AssA.
All subset results are obtained by averaging the sequence-level metrics over the selected sequences.


\begin{table}[t]
\centering
\caption{
\textbf{Performance on association-hard DanceTrack validation sequences.}
For each tracker, we select the ten sequences with the largest values of baseline $\mathrm{DetA}-\mathrm{AssA}$.
All values are averages of sequence-level metrics.
Differences are computed from unrounded values.
}
\small
\setlength{\tabcolsep}{4.0pt}
\begin{tabular}{@{}llccc@{}}
\toprule
Tracker
& Metric
& Base
& TQC
& $\Delta$ \\
\midrule
\multicolumn{5}{@{}l}{\textit{Deformable:}} \\
MOTR
& HOTA
& 51.79 & \textbf{52.77} & +0.98 \\
& AssA
& 33.79 & \textbf{34.75} & +0.96 \\
& IDF1
& 47.28 & \textbf{48.51} & +1.23 \\
\addlinespace[0.2em]
MeMOTR
& HOTA
& 47.12 & \textbf{48.68} & +1.56 \\
& AssA
& 32.37 & \textbf{34.54} & +2.17 \\
& IDF1
& 49.02 & \textbf{49.89} & +0.87 \\
\addlinespace[0.2em]
CO-MOT
& HOTA
& 56.43 & \textbf{57.91} & +1.48 \\
& AssA
& 40.99 & \textbf{43.33} & +2.34 \\
& IDF1
& 56.80 & \textbf{59.40} & +2.60 \\
\midrule
\multicolumn{5}{@{}l}{\textit{DAB:}} \\
MeMOTR
& HOTA
& 50.37 & \textbf{52.92} & +2.55 \\
& AssA
& 36.10 & \textbf{39.49} & +3.39 \\
& IDF1
& 52.90 & \textbf{56.00} & +3.10 \\
\addlinespace[0.2em]
SambaMOTR
& HOTA
& 47.67 & \textbf{49.14} & +1.47 \\
& AssA
& 32.52 & \textbf{34.51} & +1.99 \\
& IDF1
& 51.83 & \textbf{53.74} & +1.91 \\
\midrule
DualMOT$^\dagger$
& HOTA
& 57.24 & \textbf{58.36} & +1.11 \\
& AssA
& 43.44 & \textbf{44.81} & +1.37 \\
& IDF1
& 62.30 & \textbf{63.87} & +1.57 \\
\bottomrule
\end{tabular}
\label{tab:dancetrack_association_hard}
\end{table}

\begin{table}[t]
\centering
\caption{
\textbf{Performance on association-hard SportsMOT validation sequences.}
For each tracker, we select the ten sequences with the largest values of baseline $\mathrm{DetA}-\mathrm{AssA}$.
All values are averages of sequence-level metrics.
Differences are computed from unrounded values.
}
\small
\setlength{\tabcolsep}{4.0pt}
\begin{tabular}{@{}llccc@{}}
\toprule
Tracker
& Metric
& Base
& TQC
& $\Delta$ \\
\midrule
\multicolumn{5}{@{}l}{\textit{Deformable:}} \\
MeMOTR
& HOTA
& 67.00 & \textbf{68.63} & +1.63 \\
& AssA
& 52.39 & \textbf{55.04} & +2.65 \\
& IDF1
& 67.15 & \textbf{68.77} & +1.62 \\
\midrule
\multicolumn{5}{@{}l}{\textit{DAB:}} \\
MeMOTR
& HOTA
& 68.18 & \textbf{69.27} & +1.09 \\
& AssA
& 54.07 & \textbf{55.85} & +1.78 \\
& IDF1
& 68.55 & \textbf{70.13} & +1.58 \\
\addlinespace[0.2em]
SambaMOTR
& HOTA
& 70.84 & \textbf{71.31} & +0.48 \\
& AssA
& 59.00 & \textbf{59.73} & +0.73 \\
& IDF1
& 71.95 & \textbf{72.57} & +0.62 \\
\bottomrule
\end{tabular}
\label{tab:sportsmot_association_hard}
\end{table}

\subsection{Association-Hard Results on DanceTrack}
\label{sec:supp_dancetrack_association_hard}

For each tracker, we rank the DanceTrack validation sequences by baseline $\mathrm{DetA}-\mathrm{AssA}$ and select the ten sequences with the largest values.
Sequence selection uses only baseline results and is performed separately for each tracker.

Table~\ref{tab:dancetrack_association_hard} reports the results.
TQC improves HOTA, AssA, and IDF1 across all six tracker settings.
The AssA improvements range from $+0.96$ for MOTR-Deformable to $+3.39$ for MeMOTR-DAB.

\subsection{Association-Hard Results on SportsMOT}
\label{sec:supp_sportsmot_association_hard}

For each tracker, we rank the SportsMOT validation sequences by baseline $\mathrm{DetA}-\mathrm{AssA}$ and select the ten sequences with the largest values.
The ranking is performed separately for each tracker and uses only baseline results, making sequence selection independent of TQC performance.

Table~\ref{tab:sportsmot_association_hard} reports the results.
TQC improves HOTA, AssA, and IDF1 for all three tracker settings.
Among the reported metrics, the largest gain is observed in AssA for every setting.
In particular, the AssA improvements are $+2.65$ for MeMOTR-Deformable, $+1.78$ for MeMOTR-DAB, and $+0.73$ for SambaMOTR-DAB.
These results indicate that the tendency observed on DanceTrack also extends to SportsMOT.

\begin{table}[t]
\centering
\caption{
\textbf{DanceTrack results using an alternative definition of association-hard sequences.}
For each tracker, we select the ten validation sequences with the lowest baseline AssA.
All values are averages of sequence-level metrics.
Differences are computed from unrounded values.
}
\small
\setlength{\tabcolsep}{4.0pt}
\begin{tabular}{@{}llccc@{}}
\toprule
Tracker
& Metric
& Base
& TQC
& $\Delta$ \\
\midrule
\multicolumn{5}{@{}l}{\textit{Deformable:}} \\
MOTR
& HOTA
& 39.94 & \textbf{40.16} & +0.22 \\
& DetA
& 64.29 & \textbf{65.51} & +1.22 \\
& AssA
& \textbf{25.36} & 25.15 & -0.21 \\
& MOTA
& 66.83 & \textbf{68.49} & +1.67 \\
& IDF1
& 37.26 & \textbf{37.56} & +0.30 \\
\addlinespace[0.2em]
MeMOTR
& HOTA
& 44.91 & \textbf{46.72} & +1.81 \\
& DetA
& 64.63 & \textbf{65.51} & +0.87 \\
& AssA
& 31.90 & \textbf{33.92} & +2.02 \\
& MOTA
& 71.21 & \textbf{72.68} & +1.47 \\
& IDF1
& 47.95 & \textbf{49.44} & +1.49 \\
\addlinespace[0.2em]
CO-MOT
& HOTA
& 52.13 & \textbf{53.34} & +1.21 \\
& DetA
& 74.16 & \textbf{74.17} & +0.01 \\
& AssA
& 37.03 & \textbf{38.85} & +1.83 \\
& MOTA
& 82.67 & \textbf{82.74} & +0.07 \\
& IDF1
& 52.60 & \textbf{54.80} & +2.20 \\
\midrule
\multicolumn{5}{@{}l}{\textit{DAB:}} \\
MeMOTR
& HOTA
& 49.13 & \textbf{52.15} & +3.02 \\
& DetA
& 69.35 & \textbf{69.90} & +0.55 \\
& AssA
& 35.45 & \textbf{39.55} & +4.10 \\
& MOTA
& 76.53 & \textbf{77.01} & +0.48 \\
& IDF1
& 52.17 & \textbf{56.06} & +3.88 \\
\addlinespace[0.2em]
SambaMOTR
& HOTA
& 44.93 & \textbf{46.24} & +1.31 \\
& DetA
& \textbf{66.49} & 66.34 & -0.15 \\
& AssA
& 31.12 & \textbf{32.94} & +1.82 \\
& MOTA
& \textbf{74.00} & 73.84 & -0.17 \\
& IDF1
& 50.82 & \textbf{52.75} & +1.93 \\
\midrule
DualMOT$^\dagger$
& HOTA
& 53.98 & \textbf{54.79} & +0.81 \\
& DetA
& 72.86 & \textbf{73.52} & +0.66 \\
& AssA
& 40.62 & \textbf{41.33} & +0.71 \\
& MOTA
& 82.70 & \textbf{83.56} & +0.86 \\
& IDF1
& 60.14 & \textbf{60.79} & +0.65 \\
\bottomrule
\end{tabular}
\label{tab:dancetrack_lowest_assa}
\end{table}

\begin{table}[t]
\centering
\caption{
\textbf{SportsMOT results using an alternative definition of association-hard sequences.}
For each tracker, we select the ten validation sequences with the lowest baseline AssA.
All values are averages of sequence-level metrics.
Differences are computed from unrounded values.
}
\small
\setlength{\tabcolsep}{4.0pt}
\begin{tabular}{@{}llccc@{}}
\toprule
Tracker
& Metric
& Base
& TQC
& $\Delta$ \\
\midrule
\multicolumn{5}{@{}l}{\textit{Deformable:}} \\
MeMOTR
& HOTA
& 65.66 & \textbf{68.12} & +2.46 \\
& DetA
& 84.29 & \textbf{84.56} & +0.27 \\
& AssA
& 51.33 & \textbf{55.28} & +3.96 \\
& MOTA
& 93.10 & \textbf{93.52} & +0.42 \\
& IDF1
& 66.29 & \textbf{69.16} & +2.87 \\
\midrule
\multicolumn{5}{@{}l}{\textit{DAB:}} \\
MeMOTR
& HOTA
& 66.00 & \textbf{66.40} & +0.39 \\
& DetA
& 82.68 & \textbf{83.06} & +0.38 \\
& AssA
& 52.86 & \textbf{53.30} & +0.44 \\
& MOTA
& 91.53 & \textbf{91.99} & +0.46 \\
& IDF1
& 66.99 & \textbf{67.78} & +0.79 \\
\addlinespace[0.2em]
SambaMOTR
& HOTA
& 70.84 & \textbf{71.31} & +0.48 \\
& DetA
& 85.28 & \textbf{85.36} & +0.08 \\
& AssA
& 59.00 & \textbf{59.73} & +0.73 \\
& MOTA
& 93.45 & \textbf{93.68} & +0.23 \\
& IDF1
& 71.95 & \textbf{72.57} & +0.62 \\
\bottomrule
\end{tabular}
\label{tab:sportsmot_lowest_assa}
\end{table}

\subsection{Alternative Sequence Selection Using Baseline AssA}
\label{sec:supp_lowest_assa}

We next examine whether the association-hard results depend on the sequence-selection criterion.
As an alternative, we select the ten validation sequences with the lowest baseline AssA for each tracker.
This criterion directly identifies sequences with low association quality without considering their detection quality.
As in the preceding analyses, sequence selection is performed separately for each tracker and uses only baseline results, making it independent of TQC performance.

Tables~\ref{tab:dancetrack_lowest_assa} and~\ref{tab:sportsmot_lowest_assa} report the full results.
TQC improves HOTA and IDF1 across all nine tracker and dataset settings.
It also improves AssA in eight of the nine settings.
The only exception is MOTR-Deformable on DanceTrack, where AssA decreases slightly by $0.21$, while HOTA and IDF1 improve by $+0.22$ and $+0.30$, respectively.
For SambaMOTR-DAB on SportsMOT, the two sequence-selection criteria select the same ten sequences and therefore produce identical results.
Overall, the improvements are largely preserved under this alternative definition of association difficulty, although they are not uniform across every tracker and metric.

\begin{table}[t]
\centering
\caption{
\textbf{Hyperparameter sensitivity on the DanceTrack validation set.}
For each tracker, we vary one parameter at a time around the selected value while keeping the other parameter fixed.
Best values within each one-parameter sweep are shown in bold.
}
\small
\setlength{\tabcolsep}{3.0pt}
\begin{tabular}{@{}lcccccc@{}}
\toprule
Tracker
& Param.
& $\beta$
& $\rho$
& HOTA
& AssA
& IDF1 \\
\midrule
\multicolumn{7}{@{}l}{\textit{Deformable:}} \\
MOTR
& $\beta$
& 1.25 & 0.15
& 51.85 & \textbf{38.88} & 49.80 \\
& $\beta$
& 1.50 & 0.15
& \textbf{51.87} & 38.82 & 49.80 \\
& $\beta$
& 1.75 & 0.15
& 51.84 & 38.77 & \textbf{49.87} \\
\cmidrule(lr){2-7}
& $\rho$
& 1.50 & 0.10
& 51.20 & 38.11 & 48.94 \\
& $\rho$
& 1.50 & 0.15
& \textbf{51.87} & \textbf{38.82} & \textbf{49.80} \\
& $\rho$
& 1.50 & 0.25
& 51.31 & 38.10 & 49.32 \\
\midrule
MeMOTR
& $\beta$
& 1.00 & 0.05
& 61.98 & \textbf{53.84} & \textbf{65.21} \\
& $\beta$
& 1.25 & 0.05
& \textbf{61.99} & 53.80 & 65.03 \\
& $\beta$
& 1.50 & 0.05
& 61.95 & 53.72 & 65.20 \\
\cmidrule(lr){2-7}
& $\rho$
& 1.25 & 0.05
& \textbf{61.99} & \textbf{53.80} & \textbf{65.03} \\
& $\rho$
& 1.25 & 0.10
& 61.06 & 52.24 & 64.39 \\
\midrule
CO-MOT
& $\beta$
& 0.75 & 0.10
& 65.55 & 55.52 & 67.86 \\
& $\beta$
& 1.00 & 0.10
& \textbf{65.91} & \textbf{56.01} & \textbf{68.36} \\
& $\beta$
& 1.25 & 0.10
& 65.59 & 55.50 & 67.81 \\
\cmidrule(lr){2-7}
& $\rho$
& 1.00 & 0.05
& 65.10 & 54.90 & 67.27 \\
& $\rho$
& 1.00 & 0.10
& \textbf{65.91} & \textbf{56.01} & \textbf{68.36} \\
& $\rho$
& 1.00 & 0.15
& 65.50 & 55.58 & 67.99 \\
\midrule
\multicolumn{7}{@{}l}{\textit{DAB:}} \\
MeMOTR
& $\beta$
& 1.50 & 0.15
& 64.83 & \textbf{56.47} & 68.35 \\
& $\beta$
& 1.75 & 0.15
& \textbf{64.87} & 56.35 & \textbf{68.50} \\
& $\beta$
& 2.00 & 0.15
& 64.32 & 55.53 & 67.63 \\
\cmidrule(lr){2-7}
& $\rho$
& 1.75 & 0.10
& 64.24 & 55.67 & 67.66 \\
& $\rho$
& 1.75 & 0.15
& \textbf{64.87} & \textbf{56.35} & \textbf{68.50} \\
& $\rho$
& 1.75 & 0.20
& 64.58 & 55.83 & 67.81 \\
\midrule
SambaMOTR
& $\beta$
& 0.50 & 0.50
& 62.28 & 52.37 & 66.96 \\
& $\beta$
& 0.75 & 0.50
& \textbf{63.13} & \textbf{53.86} & \textbf{67.65} \\
& $\beta$
& 1.00 & 0.50
& 62.40 & 52.70 & 67.04 \\
\cmidrule(lr){2-7}
& $\rho$
& 0.75 & 0.25
& 62.50 & 52.70 & 66.53 \\
& $\rho$
& 0.75 & 0.50
& \textbf{63.13} & \textbf{53.86} & \textbf{67.65} \\
& $\rho$
& 0.75 & 0.75
& 62.95 & 53.55 & 67.58 \\
\midrule
DualMOT$^\dagger$
& $\beta$
& 1.75 & 0.25
& 67.97 & 59.44 & 73.01 \\
& $\beta$
& 2.00 & 0.25
& \textbf{68.07} & 59.49 & \textbf{73.03} \\
& $\beta$
& 2.25 & 0.25
& 68.06 & \textbf{59.59} & 73.00 \\
\cmidrule(lr){2-7}
& $\rho$
& 2.00 & 0.15
& 67.64 & 59.21 & 72.84 \\
& $\rho$
& 2.00 & 0.25
& \textbf{68.07} & \textbf{59.49} & \textbf{73.03} \\
& $\rho$
& 2.00 & 0.50
& 67.54 & 58.84 & 72.04 \\
\bottomrule
\end{tabular}
\label{tab:hyperparam_sensitivity_dancetrack_val}
\end{table}

\begin{table}[t]
\centering
\caption{
\textbf{Hyperparameter sensitivity on the SportsMOT validation set.}
For each tracker, we vary one parameter at a time around the selected value while keeping the other parameter fixed.
Best values within each one-parameter sweep are shown in bold.
}
\small
\setlength{\tabcolsep}{3.0pt}
\begin{tabular}{@{}lcccccc@{}}
\toprule
Tracker
& Param.
& $\beta$
& $\rho$
& HOTA
& AssA
& IDF1 \\
\midrule
\multicolumn{7}{@{}l}{\textit{Deformable:}} \\
MeMOTR
& $\beta$
& 1.25 & 0.15
& 75.48 & 66.66 & 78.26 \\
& $\beta$
& 1.50 & 0.15
& \textbf{75.65} & 66.92 & \textbf{78.48} \\
& $\beta$
& 1.75 & 0.15
& 75.64 & \textbf{66.93} & 78.46 \\
\cmidrule(lr){2-7}
& $\rho$
& 1.50 & 0.10
& 75.48 & 66.62 & 78.22 \\
& $\rho$
& 1.50 & 0.15
& \textbf{75.65} & \textbf{66.92} & \textbf{78.48} \\
& $\rho$
& 1.50 & 0.25
& 75.25 & 66.33 & 78.07 \\
\midrule
\multicolumn{7}{@{}l}{\textit{DAB:}} \\
MeMOTR
& $\beta$
& 0.50 & 0.10
& 76.24 & 67.43 & 78.93 \\
& $\beta$
& 0.75 & 0.10
& \textbf{76.87} & \textbf{68.44} & \textbf{79.82} \\
& $\beta$
& 1.00 & 0.10
& 76.68 & 68.10 & 79.53 \\
\cmidrule(lr){2-7}
& $\rho$
& 0.75 & 0.05
& 75.97 & 66.97 & 78.62 \\
& $\rho$
& 0.75 & 0.10
& \textbf{76.87} & \textbf{68.44} & \textbf{79.82} \\
& $\rho$
& 0.75 & 0.15
& 76.43 & 67.68 & 79.34 \\
\midrule
SambaMOTR
& $\beta$
& 0.25 & 0.75
& 78.46 & 71.40 & 82.61 \\
& $\beta$
& 0.50 & 0.75
& \textbf{78.79} & \textbf{71.96} & \textbf{82.98} \\
& $\beta$
& 0.75 & 0.75
& 78.76 & 71.95 & 82.87 \\
\cmidrule(lr){2-7}
& $\rho$
& 0.50 & 0.25
& 78.78 & \textbf{71.96} & \textbf{82.99} \\
& $\rho$
& 0.50 & 0.50
& \textbf{78.79} & \textbf{71.96} & 82.98 \\
& $\rho$
& 0.50 & 0.75
& \textbf{78.79} & \textbf{71.96} & 82.98 \\
\bottomrule
\end{tabular}
\label{tab:hyperparam_sensitivity_sportsmot_val}
\end{table}

\section{Hyperparameter Sensitivity}
\label{sec:supp_hyperparameter_sensitivity}

We examine the sensitivity of TQC to the correction parameters $\beta$ and $\rho$ on the DanceTrack and SportsMOT validation sets.
Tables~\ref{tab:hyperparam_sensitivity_dancetrack_val} and~\ref{tab:hyperparam_sensitivity_sportsmot_val} report the results of varying one parameter at a time while keeping the other fixed.
The results show that the degree of sensitivity varies across tracker and dataset settings.

\begin{table*}[t]
\centering
\caption{
\textbf{Cross-dataset transfer of TQC hyperparameters on the test sets.}
DT and SP denote DanceTrack and SportsMOT, respectively.
Transferred parameters are selected on the source validation set, whereas target-specific parameters are selected on the target validation set.
}
\small
\setlength{\tabcolsep}{6.0pt}
\begin{tabular}{@{}llccc@{}}
\toprule
Tracker
& Transfer
& $(\beta,\rho)$
& \shortstack{Transferred\\$\Delta$HOTA}
& \shortstack{Target-specific\\$\Delta$HOTA} \\
\midrule
MeMOTR-Deformable
& DT $\rightarrow$ SP
& (1.25, 0.05)
& +0.10 & +0.41 \\
MeMOTR-DAB
& DT $\rightarrow$ SP
& (1.75, 0.15)
& +0.10 & +0.14 \\
SambaMOTR-DAB
& DT $\rightarrow$ SP
& (0.75, 0.50)
& +0.19 & +0.08 \\
\midrule
MeMOTR-Deformable
& SP $\rightarrow$ DT
& (1.50, 0.15)
& +0.88 & +0.91 \\
MeMOTR-DAB
& SP $\rightarrow$ DT
& (0.75, 0.10)
& -0.87 & +0.41 \\
SambaMOTR-DAB
& SP $\rightarrow$ DT
& (0.50, 0.75)
& +0.48 & +0.32 \\
\midrule
Average
& &
& +0.15 & +0.38 \\
\bottomrule
\end{tabular}
\label{tab:supp_hyperparameter_transfer}
\end{table*}

\section{Cross-Dataset Hyperparameter Transfer}
\label{sec:supp_hyperparameter_transfer}

We examine whether TQC retains its improvements when the correction parameters are selected on a different dataset.
For each tracker, we select $(\beta,\rho)$ using validation HOTA on the source dataset and apply them to the target test set without further tuning.
The pretrained tracker checkpoint used for the target dataset remains unchanged, and only the correction parameters are transferred.

As shown in Table~\ref{tab:supp_hyperparameter_transfer}, TQC retains positive HOTA gains in five of the six test settings without selecting parameters on the target validation set.
The average gain is $+0.15$, compared with $+0.38$ using target-specific selection.
These results show that improvements can transfer across datasets, although the effectiveness of the selected parameters varies across settings.

\begin{table}[t]
\centering
\caption{
\textbf{Test results on MOT17 and MOT20 with MeMOTR-DAB.}
Both evaluations use the official MOT17 checkpoint without additional training on MOT20.
TQC uses $(\beta,\rho)=(0.75,0.25)$, selected on MOT20 train and fixed for both test sets.
}
\small
\setlength{\tabcolsep}{3.5pt}
\begin{tabular}{@{}llccccc@{}}
\toprule
Dataset & Method & HOTA & DetA & AssA & MOTA & IDF1 \\
\midrule
MOT17
& MeMOTR-DAB
& 58.8 & 59.6 & \textbf{58.4} & 72.9 & 71.5 \\
& \quad + TQC
& 58.8 & \textbf{59.7} & 58.3
& \textbf{73.0} & \textbf{71.7} \\
\midrule
MOT20
& MeMOTR-DAB
& 36.8 & 33.7 & \textbf{40.4} & 40.6 & 46.0 \\
& \quad + TQC
& \textbf{37.5} & \textbf{35.1} & 40.2
& \textbf{42.6} & \textbf{47.4} \\
\bottomrule
\end{tabular}
\label{tab:supp_mot_test}
\end{table}

\begin{table}[t]
\centering
\caption{
\textbf{Results on BDD100K validation.}
All metrics are averaged over eight classes.
mAssocA and mAssA are association metrics from TETA and HOTA, respectively.
}
\small
\setlength{\tabcolsep}{2.5pt}
\begin{tabular}{@{}lccccc@{}}
\toprule
Method & mTETA & mHOTA & mLocA & mAssocA & mAssA \\
\midrule
MeMOTR-DAB
& 52.85 & 40.59 & \textbf{39.63} & 56.05 & 52.74 \\
\quad + TQC
& \textbf{52.90} & \textbf{40.64} & 39.62
& \textbf{56.28} & \textbf{52.78} \\
\bottomrule
\end{tabular}
\label{tab:supp_bdd100k}
\end{table}

\section{Additional Dataset Evaluations}
\label{sec:supp_additional_datasets}

\subsection{MOT17 and MOT20}

We further evaluate TQC on MOT17~\cite{milan2016mot16} and MOT20~\cite{dendorfer2020mot20} using MeMOTR-DAB with the official MOT17 checkpoint.
The tracker is not retrained on MOT20.
Since the official checkpoint was trained on MOT17 train, we select $(\beta,\rho)=(0.75,0.25)$ using MOT20 train and keep these parameters fixed for both test sets.

Table~\ref{tab:supp_mot_test} reports test results on MOT17 and MOT20.
On MOT17, HOTA remains $58.8$ at the reported precision, with slight improvements in DetA, MOTA, and IDF1 and a slight decrease in AssA.
On MOT20, TQC improves HOTA from $36.8$ to $37.5$ and IDF1 from $46.0$ to $47.4$.
The improvement is accompanied by higher DetA ($+1.4$) and MOTA ($+2.0$), while AssA decreases slightly by $0.2$.
IDSWs also decrease from $1{,}056$ to $1{,}047$ on MOT20.

\subsection{BDD100K}

We further evaluate TQC using the official MeMOTR-DAB checkpoint for BDD100K~\cite{yu2020bdd100k} on its validation set of 200 videos.
We use $K=5$ and the parameters $(\beta,\rho)=(0.75,0.10)$ selected on SportsMOT validation for MeMOTR-DAB, without further tuning on BDD100K.
Candidate tracks are selected across all eight object classes.

As shown in Table~\ref{tab:supp_bdd100k}, TQC slightly improves mTETA by $0.05$, mHOTA by $0.05$, mAssocA by $0.23$, and mAssA by $0.04$, while mLocA remains nearly unchanged.
These results indicate that TQC can be applied to multi-class tracking without dataset-specific parameter tuning, although the performance gains are limited.

\begin{table}[t]
\centering
\caption{
\textbf{Computational complexity of TQC.}
Base denotes the estimated computation of the tracker forward pass, and TQC denotes the additional calibration cost.
We follow a common profiling convention that counts each multiply-accumulate as one operation.
Here, \#Q denotes the number of active track-query slots, and Params denotes the total number of model parameters.
TQC introduces no additional learnable parameters.
}
\small
\setlength{\tabcolsep}{4pt}
\begin{tabular}{@{}lcccc@{}}
\toprule
Tracker
& \#Q
& Params
& \shortstack{Base\\(GFLOPs)}
& \shortstack{TQC\\(MFLOPs)} \\
\midrule
\multicolumn{5}{@{}l}{\textit{Deformable:}} \\
MOTR
& 2
& 44.14M
& 233.28
& 0.0114 \\
& 23
& 44.14M
& 233.57
& 1.1305 \\
\addlinespace[0.2em]
MeMOTR
& 2
& 42.88M
& 233.28
& 0.0114 \\
& 23
& 42.88M
& 233.50
& 1.1275 \\
\addlinespace[0.2em]
CO-MOT
& 6
& 41.94M
& 232.04
& 0.0836 \\
& 69
& 41.94M
& 232.71
& 9.9368 \\
\midrule
\multicolumn{5}{@{}l}{\textit{DAB:}} \\
MeMOTR
& 2
& 50.36M
& 310.19
& 0.0114 \\
& 23
& 50.36M
& 310.53
& 1.1275 \\
\addlinespace[0.2em]
SambaMOTR
& 2
& 49.34M
& 310.21
& 0.0114 \\
& 23
& 49.34M
& 310.82
& 1.1275 \\
\midrule
DualMOT
& 2
& 48.12M
& 327.44
& 0.0114 \\
& 23
& 48.12M
& 328.98
& 1.1275 \\
\bottomrule
\end{tabular}
\label{tab:supp_model_complexity}
\end{table}

\section{Computational Complexity}
\label{sec:supp_complexity}

We further analyze the computational complexity of TQC.
All estimates use batch size 1 and an input resolution of $800 \times 1440$.
We profile six tracker configurations under a common counting protocol, using 300 detection queries for MOTR, MeMOTR, SambaMOTR, and DualMOT, and 180 detection-query slots for CO-MOT.

We evaluate two settings with 2 and 23 target identities.
The 2-identity setting is the smallest case in which correction between distinct tracks is possible.
The 23-identity setting corresponds to the maximum number of valid pedestrians per frame observed in the DanceTrack validation set.
CO-MOT uses three shadow queries per identity, resulting in 6 and 69 active query slots, respectively.
For DualMOT, we account for the historical query features used by its temporal decoder, assuming a fully populated memory bank of size three.

The additional computation introduced by TQC is negligible compared with the full tracker forward pass.
Across the profiled settings, TQC adds at most $9.94$ MFLOPs, while the baseline trackers require at least $232.04$ GFLOPs.

\FloatBarrier

\end{document}